\documentclass[runningheads]{llncs}
\usepackage[table,xcdraw,dvipsnames]{xcolor}
 
\usepackage{eccv}

\usepackage{eccvabbrv}

\usepackage{graphicx}
\usepackage{booktabs}
\usepackage[accsupp]{axessibility}  

\usepackage{hyperref}

\usepackage{orcidlink}

\usepackage{multirow, adjustbox}
\usepackage{soul,color}
\DeclareMathOperator{\pcd}{\mathcal{P}}

\DeclareMathOperator*{\argmin}{argmin}

\newcommand{\hide}[1]{}

\begin{document}

\title{
SelfGeo: Self-supervised and Geodesic-consistent Estimation of  Keypoints on Deformable Shapes
} 

\titlerunning{SelfGeo}

\author{Mohammad Zohaib\inst{1}\orcidlink{0000-0003-2259-4121} \and
Luca Cosmo\inst{2}\orcidlink{0000-0001-7729-4666} \and
Alessio {Del Bue}\inst{1}\orcidlink{0000-0002-2262-4872} }

\authorrunning{M.~Zohaib et al.}

\institute{Pattern Analysis \& Computer Vision, 
Italian Institute of Technology, Genoa, Italy \and
Ca' Foscari University of Venice, Venice, Italy
\\
\email{zohaib.mohammad@hotmail.com, luca.cosmo@unive.it, alessio.delbue@iit.it}
}

\maketitle

\begin{abstract}
Unsupervised 3D keypoints estimation from Point Cloud Data (PCD) is a complex task, even more challenging when an object shape is deforming. 
As keypoints should be semantically and geometrically consistent across all the 3D frames -- each keypoint should be anchored to a specific part of the deforming shape irrespective of intrinsic and extrinsic motion. 
This paper presents, ``SelfGeo'', a self-supervised method that computes persistent 3D keypoints of non-rigid objects from arbitrary PCDs without the need of human annotations. 
The gist of SelfGeo is to estimate keypoints between frames that respect invariant properties of deforming bodies. Our main contribution is to enforce that keypoints deform along with the shape while keeping constant geodesic distances among them. 
This principle is then propagated to the design of a set of losses which minimization let emerge repeatable keypoints in specific semantic locations of the non-rigid shape.
We show experimentally that the use of geodesic has a clear advantage in challenging dynamic scenes and with different classes of deforming shapes (humans and animals).
Code and data are available at: \href{https://github.com/IIT-PAVIS/SelfGeo}{https://github.com/IIT-PAVIS/SelfGeo}
%
%
\end{abstract}

\section{Introduction}
\label{sec:intro}
\begin{figure}[t]
  \centering
    \includegraphics[width=.75\linewidth]{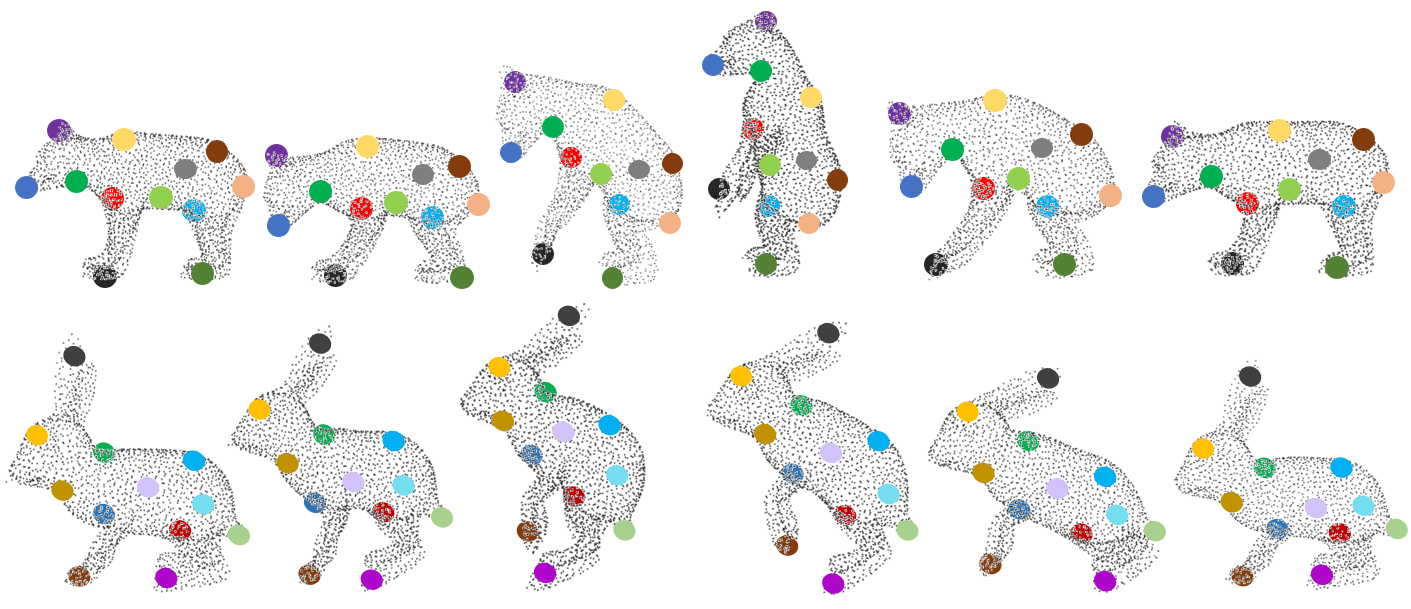}
  \caption{Overview of the \textit{SelfGeo}. 
  The keypoints estimated for the two shapes of the same category are temporally consistent; they are anchored to their locations (equal geodesic distances) between two frames regardless of deformation, and maintain the semantic information (same colours indicate the corresponding keypoints).
  Moreover, the keypoints are estimated close to the surface and covering the whole shape. 
  }
  \label{fig:overview}
\end{figure}

Modelling and representing 3D deformable/non-rigid shapes is fundamental for enabling Augmented Reality, Virtual Reality and Human-Robot Interaction applications in realistic dynamic environments.  
In particular, the skeleton of a shape plays a vital role by providing information about the object's structure, poses and actions~\cite{ref:tang2022lake,ref:fernandez2020unsupervised,ref:zhong20233d,ref:su2021nerf,ref:dai2023animal}. 
To generate a skeleton of a shape, the common practice is to estimate a set of keypoints~\cite{ref:shi2021skeleton} such that they are temporally consistent and highlight specific semantic parts of the object.
However, estimating such keypoints for deformable shapes is a very challenging task due to their intrinsic deformations and intra-class variations.

Some existing approaches compute keypoints in a partially/fully supervised way, where high-quality ground truth annotations are requested~\cite{ref:weng20233d,ref:zanfir2023hum3dil}.
Crucially, labelling points on deformable shapes is a daunting, time and resource-consuming task that is also prone to human errors. Most current methods are limited to synthetic datasets where correspondences through frames are given by construction and no annotated keypoints currently exist for real datasets. 

Recently, Weng et al.~\cite{ref:weng20233d} estimated 3D keypoints for a synthetic human shape by using existing ground truth keypoints and later refining them with unsupervised losses to increase generalization to the real data. Unfortunately, due to the nature of the used loss functions, this approach is limited to human shapes.
Differently, Fernandez et al.~\cite{ref:fernandez2020unsupervised}  use a linear shape basis for the ordered point correspondences.
They learn to estimate consistent 3D keypoints by considering the objects' symmetry. Therefore, they perform very well for symmetric objects such as chairs, airplanes, etc. However, deforming motion may violate symmetry (i.e. having only one arm upright), thus drastically reducing the performance of the approach.
Recent literature dealing with 3D deformable shapes \cite{ref:sengupta2024totem,ref:huang2024skeleton,ref:cosmo20223d,ref:saleh2022bending,ref:maharjan2022registration,ref:yang2021continuous,ref:tan2021humangps} has shown that injecting an isometric prior into Neural Networks architectures, either in the form of geometric losses or specifically designed components, is a good strategy to ease the learning task on the complex deformation patterns typical of deformable shapes.

Following this principle, we propose a novel self-supervised approach, named \textit{SelfGeo}, to estimate stable keypoints that implicitly distil geometrical and temporal properties of non-rigid shapes.  
Unlike~\cite{ref:zhong20233d, ref:fernandez2020unsupervised}, during training, we take a sequence of PCDs as input and estimate a set of keypoints for each input PCD. These keypoints should satisfy determined properties that we promote by a set of differentiable losses. The \textbf{Shape loss} acts on the distribution of keypoints on a single PCD, and it is mainly adapted from previous works on rigid keypoints estimation \cite{ref:zohaib2023sc3k,ref:fernandez2020unsupervised,ref:zohaib20223d,ref:suwajanakorn2018discovery,ref:zohaib4349267cdhn}. It enforces that the keypoints cover the volume of the object, 
adhere to its surface (i.e. compactness) and are informative enough to reconstruct the full PCD. 
Our main novelty is the introduction of a \textbf{Deformation loss}.
It accounts for the topological structure of the deforming body by enforcing that the geodesic distances do not change among the keypoints estimated for all the PCDs of the input sequence. To further impose temporal continuity during deformation, we also introduce a pairwise regularisation in consecutive frames. Our architecture shows that, at test time, keypoints emerge with the desirable properties we are seeking (Fig.~\ref{fig:overview}): they are localised in semantically meaningful regions while covering the whole surface and they are anchored to their locations regardless non-rigid and rigid motion of the deformable shape.

To summarise, the main contributions of this work are as follows: 
\begin{itemize}
    \item We propose a self-supervised approach to estimate 3D keypoints from PCDs that does not require any a priori information about the deforming shape.
    \item To this end, we propose a novel 
    differentiable loss function that preserves the geodesic distances between keypoints.
    \item We evaluate the CAPE and Deforming Things 4D dataset showing that the proposed approach is general and can be used for any deformable shapes irrespective of body structure or skeleton.
    \item The evaluation on the real ITOP dataset and the 
    ablation studies show that the proposed approach is robust to noisy or decimated PCDs. 
\end{itemize}
    
The paper is organized as follows. Section 2 reports the existing methods, Section 3 proposes the approach, Section 4 presents the experiments and discussions on the results, and finally, the conclusions are given in Section 5.
\section{Related work}
\label{sec:literature_review}
3D keypoints are estimated from images or point clouds for different downstream tasks such as finding distinct locations on objects~\cite{ref:bai2023coke,ref:gupta2023learning,ref:shi2023optimal}, object deformation~\cite{ref:jakab2021keypointdeformer,ref:yang2022object,ref:wang2022keypoint}, generalizable manipulation~\cite{ref:xue2023useek}, shape matching~\cite{ref:attaiki2022ncp,ref:attaiki2023generalizable,ref:kim2023semantic}, clothes perception (folding, laundry, and
dressing)~\cite{ref:zhou2023clothesnet}, etc. Based on the object's geometry, we cluster existing methods into two sections: rigid or non-rigid. 
\subsection{Keypoints estimation for rigid objects}
Suwajanakorn et al.~\cite{ref:suwajanakorn2018discovery} present an approach to estimate 3D keypoints in the form of 2D positions and depth from a pair of images. Their approach forces 2D keypoints to be estimated within the object silhouette and uses known camera projections.
A similar approach is proposed by Zohaib et al., in~\cite{ref:zohaib20223d} that estimates 3D keypoints directly from images. During the training, they utilize the 3D ground truths to localize the kepoints. The presented results validate that their approach overperforms Suwajanakorn's method~\cite{ref:suwajanakorn2018discovery} when estimating the object pose. 
The SC3K approach~\cite{ref:zohaib2023sc3k} estimates semantically coherent keypoints for rigid objects from point clouds in a self-supervised way. This approach is robust to the pose, down-sampling and noise in the input PCDs. 
A similar approach is presented by Li et al.~\cite{ref:li2019usip} that first generates clusters from the input point clouds and then estimates a keypoint for every cluster. Their method requires two rotated versions of a PCD for learning the stable keypoints under arbitrary transformations.
Xue et al.~\cite{ref:xue2023useek} presents a teacher-student structure
to discover SE(3)-equivariant keypoints estimation from point clouds.
The teacher module is similar to Skeleton Merger~\cite{ref:shi2021skeleton}, allowing keypoints estimation in the canonical pose. Whereas, the student module uses a SPRIN backbone~\cite{ref:you2021prin} to estimate the same keypoints from an object in a random pose. 
Keypoints can also be used for shape completion tasks as presented by Tang et al. in~\cite{ref:tang2022lake}.
Their approach, ``LAKe-Net'', first estimates aligned keypoints, and then uses them to generate a surface-skeleton that represents the object's topological information. The skeleton is later used in the shape reconstruction and completion.
Another method is proposed by Yuan et al. in~\cite{ref:yuan2022unsupervised} that estimates keypoints from one object that are good for reconstructing any instance of the same category. During training, their method estimates two sets of 3D semantically consistent keypoints from two different objects of the same category and uses them for self/mutual reconstruction. 
Shi et al.~\cite{ref:shi2023optimal} uses the keypoints in shape and pose estimation in robotics applications, e.g. autonomous driving. They estimate 2D/3D keypoints from the RGB/RGBD images and use the proposed graph-theoretic framework (ROBIN) to remove outliers. Finally, the resulting keypoints are used to obtain the pose and shape of an object. 

The techniques adopted in~\cite{ref:suwajanakorn2018discovery,ref:zohaib2023sc3k,ref:li2019usip,ref:zohaib20223d,ref:shi2021skeleton} to localize keypoints w.r.t. to an object's shape are significant in the self-supervised settings. Albeit they have used them for rigid objects, we use similar ideas in designing one of  our losses that helps our network to estimate keypoints covering the deformable object~\cite{ref:suwajanakorn2018discovery,ref:zohaib20223d} and adhere to the surface~\cite{ref:shi2021skeleton,ref:zohaib2023sc3k}.

\subsection{Keypoints estimation for deformable objects}
Estimating the keypoints for deformable objects, such as humans and animals, is challenging due to the irregular deformation/motion of the object's parts. 
Chen et al. in~\cite{ref:chen2021unsupervised} estimate 3D keypoints to represent robot joints and to capture an object's motion. 
Their approach uses multi-view images in an unsupervised way to generate 3D keypoints in the form of 2D heat maps and depths [\textit{u,v,d}] for each image. 
These keypoints are aggregated using the camera parameters to obtain the global coordinates of each keypoint. 
Zhong et al.~\cite{ref:zhong2022snake} presents a self-supervised approach (SNAKE) that 
jointly reconstructs the object's surface and estimates the 3D keypoints. 
The method estimates keypoint saliency for each continuous query point instead of a discrete input point cloud.
Fernandez et al.~\cite{ref:fernandez2020unsupervised}, present an unsupervised 3D keypoints estimation method that assumes object's being symmetric. The network estimates \textit{N} keypoints and applies nonmax-suppression for selecting the final keypoints. Their main focus is on rigid objects but the presented results show that their approach is also valid for the deformable human body.
An approach UKPGAN is presented in~\cite{ref:you2020ukpgan} that estimates the keypoints and semantic embeddings by optimizing the reconstruction task. Similarly to~\cite{ref:fernandez2020unsupervised}, this method is also used for human bodies in different non-rigid deformation.
Weng et al.~\cite{ref:weng20233d} propose an approach (GC-KPL) that estimates 3D keypoints from the point clouds of the human body. It is first trained on a synthetic dataset~\cite{ref:loper2015smpl} using the available ground truths to learn keypoints' positions and semantic segmentation to localize the object's part. Then, the approach is refined on the Waymo Open Dataset~\cite{ref:sun2020scalability} using an un-supervised loss functions. 
In contrast, 
Zhong et al.~\cite{ref:zhong20233d} estimate self-supervised 3D keypoints mainly for robotic manipulation tasks.
Their approach estimates 3D keypoints for two versions of the same PCD of an articulated object in different poses. The keypoints are then used to compute the rotation axis between the two PCDs. The available rigid transformations are given as a prior for network optimization. 
The method also reconstructs the target shape from the source by transporting the features of the target keypoints to the features of the source shape. 

The above methods are either supervised or are object-specific (i.e. skeleton-based and applicable only for humans) and they fail to produce consistent keypoints in case of skeleton-free deformation (i.e. for both humans and animals). In comparison, we propose \textit{SelfGeo}, a self-supervised approach that estimates 3D keypoints for any deformable objects. This is achieved by designing a deformable loss that preserves both temporal and geometric consistency among the keypoints estimated for a sequence of frames.

\section{Proposed approach}
\label{sec:proposed_approach}
\begin{figure*}[t]
  \centering
    \includegraphics[width=.90\linewidth]{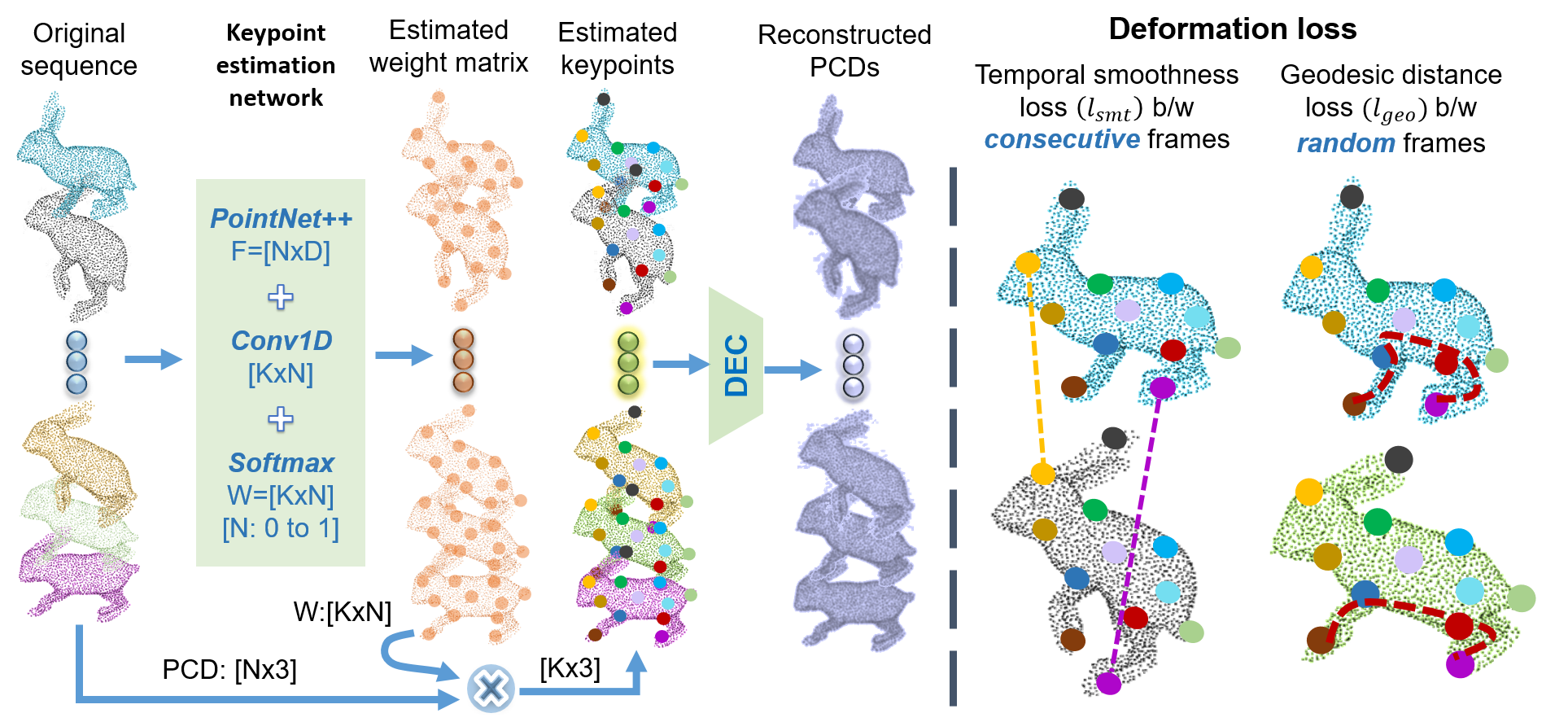}
    \caption{
    Left: proposed \textit{SelfGeo}. A sequence of PCD is input one by one to the Keypoints estimation network, which contains a PointNet++ encoder, a Conv1D and a Softmax layer. The network generates $K$ values for each point, indicating its probability to be one of the $K$ keypoints. The expected keypoint positions for each PCD are computed and 
    passed to a decoder (DEC), which consists of 4 Con1D layers, to reconstruct the 3D shape.
    To improve the keypoints' inference, \textit{SelfGeo} computes the shape loss (reconstruction, coverage and surface loss) using a single PCD, and deformation loss (right: geodesic distance and temporal smoothing loss) between two frames.}
    \label{fig:architecture}
\end{figure*}

This section describes the design of the \textit{SelfGeo}, 
the loss functions that are used to localize the keypoints, and the complete pipeline for training and testing.  

Given a PCD $\pcd = [x_1, x_2, ..., x_N] \in \mathbb{R}^{3 \times N}$ with $N$ points, the goal of the proposed approach is to estimate a set of $K$ keypoints  $\mathcal{K} = \{k_{1}, k_{2}, ..., k_{K}\} \in \pcd$ located at specific regions, 
robust to the non-rigid deforming motion, and consistent among different subjects.
To achieve this goal, we train a neural network to predict a set of $K$ probability distributions $p_{k_j}(x_i)\; \forall\; k_j \in \mathcal{K}$, representing the probability of each $x_i \in \pcd$ to be the $j^{th}$ keypoint. Then,  the position of a keypoint $k_j$ is computed as a convex combination of all $x_i \in \pcd$ as $\mathbb{E}_{\pcd}[k_j] = \sum_i x_i \:\: p_{k_j}(x_i)$.

\subsection{Self-supervised keypoints estimation losses}
We design \textit{SelfGeo} to extract $K$ keypoints which are informative, semantically consistent, well spread over the PCD and robust to non-rigid motion.
To ensure these properties, we devise two types of self-supervised losses. The {\bf shape loss} ($\mathcal{L}_{sha}$) is tasked to promote the extracted keypoints to be as much informative as possible, and ensures that they are well distributed over the PCD. This loss acts on each PCD individually and does not account for the deformation dynamics over time. The {\bf deformation loss} ($\mathcal{L}_{def}$), on the other hand, takes care of the consistency of the keypoints given the non-rigid motion of the PCDs.

The overall training loss is then 
the sum of the localization and deformation losses:
\begin{equation}
  \mathcal{L}_{tot} = \mathcal{L}_{sha} + \mathcal{L}_{def}.
 \label{eq:train_loss} 
\end{equation}

\subsubsection{Shape loss}
The shape loss is made up of three 
components: reconstruction loss, coverage loss, and surface loss. All these losses apply to a single PCD and they enforce a distribution of keypoints on the shape that respects different principles.

\paragraph{Reconstruction loss.}
Without having the access to any labeled data, it is not trivial to define what ``good'' keypoints are.
Ideally, we would like our keypoints to capture as much information of the original shape as possible. To this end, we pair our keypoints estimation with a proxy PCD reconstruction task.
The expected locations of all keypoints $\mathbb{E}_{\pcd}[k_j]$, 
are fed to a decoder (DEC) tasked to output a 3D PCD with $M$ points $\tilde{\pcd}$ = DEC($\{\mathbb{E}_{\pcd}[k_j]\}_{j=1}^K$) $= [\tilde{x}_1,\tilde{x}_2,\dots,\tilde{x}_M]$ minimizing a reconstruction loss in terms of Chamfer distance from the input point cloud $\pcd$:

\begin{equation}
  \mathcal{L}_{rec} = \sum_{i=1}^N \min_{j} \big\| x_i - \tilde{x}_j\big\|_2^2 + 
                          \sum_{j=1}^M \min_{i} \big\| \tilde{x}_j - x_i\big\|_2^2.
 \label{eq:reconstruction} 
\end{equation}
Notice here that we are flexible in the number of input $N$ and output $M$ points.
\hide{Note that, the expected location is not necessarily on the surface.}

\paragraph{Coverage loss.} One problem of the reconstruction loss is that it tends to focus keypoints positions on regions that undergo major deformations, while leaving uncovered more ``static'' regions (e.g. the head in a human is unlikely to attract any keypoints). On the other hand, we would like the estimated keypoints to highlight different parts of an object. We promote this behaviour by penalizing keypoints for being too close to each other:
\begin{equation}
    \mathcal{L}_{cov} =   \left( \frac{1}{K} \sum\limits_{i=1}^{K} \min_{j \neq i}\big\| \mathbb{E}_{\pcd} [k_i] - \mathbb{E}_{\pcd} [k_j] \big\|_2  + \epsilon \right)^{-1},
 \label{eq:separation_loss} 
\end{equation}
where epsilon is a small constant (1e-2) to prevent numeric errors.

\paragraph{Surface loss.}
Being the expected location of a keypoint $k_j$ expressed as a convex combination of all $x_i \in \pcd$, its position might be located far from the surface of the shape. This behaviour is further accentuated by $\mathcal{L}_{cov}$ whose goal is to push keypoints far apart in the Euclidean space. This would result in probability distributions of keypoints being spread over regions external to the input PCD. 
To address this issue, we introduce the surface loss ($\mathcal{L}_{surf}$), which restricts the keypoints to be estimated close to the PCD. 
This loss minimizes the expected distance of a keypoint $k_i$ in $\mathcal{K}$ from the nearest point $x \in \mathcal{P}$ as:

\begin{equation}
  \mathcal{L}_{surf} = \frac{1}{K} \sum_{i=1}^{K} \min_j \big\| \mathbb{E}_{\pcd}[k_i] - x_j \big\|_2.
 \label{eq:shape_loss} 
\end{equation}
The effect of this loss is to promote local support of the keypoints' probabilities.

The total shape loss can be summarized as a weighted sum of the above components as:
\begin{equation}
  \begin{aligned}
\mathcal{L}_{sha} = \lambda_{rec} \cdot \mathcal{L}_{rec}+ \lambda_{cov} \cdot \mathcal{L}_{cov} +  \lambda_{surf} \cdot \mathcal{L}_{surf},
  \end{aligned}
\label{eq:single-frame} 
\end{equation}
where \{$\lambda_{rec}$, $\lambda_{cov}$, $\lambda_{surf}$\} are 
experimentally set to \{1, 2.5, 6\}, respectively. 

\subsubsection{Deformation loss}
The shape loss, albeit allowing the network to identify the most informative points in the PCD, does not impose any spatial consistency of the extracted keypoints between different non-rigid poses, resulting in the keypoints drifting along the surface and swapping in position. 
To prevent this effect, we devise a deformation loss which takes into account the behavior of the keypoints on a sequence of deforming PCDs.

At training time, we assume to have access to a sequence of $T$ consecutive PCDs $[\pcd^1,\dots,\pcd^T]$ of a non-rigid body under relative motion 
with points $\pcd^t = [x^t_1, x^t_2, ..., x^t_N]$. 
A desirable property of the estimated keypoints is that they should move relatively to the deformation field while keeping their position over the object surface, thus preserving their semantic position and order against the non-rigid motion. 
Considering this principle, we propose a deformation loss based on two components: geodesic and smoothing losses. 

\paragraph{Geodesic loss.} 
Differently from a rigid setting, where only global rigid transformations (i.e. roto-translations) are considered~\cite{ref:li2019usip,ref:zohaib2023sc3k}, in the non-rigid setting the individual parts forming a PCD may change their relative positions due to deformations~\cite{ref:cosmo2020limp,ref:halimi2019unsupervised}.
This implies that, while Geodesic distances on the underlying surface are still (approx.) preserved, this is not true for the Euclidean ones.

Considering this, we present a novel differentiable loss function to preserve the expected geodesic distances between the estimated keypoints along all frames.
Given the PCD $\pcd^t$ at frame $t$, the expected geodesic distance between two keypoints is defined as:
\begin{equation*}
    \mathbb{E}_{\pcd^t}[d_{\pcd^t}(k_i,k_j)] = \sum_{m,n} p_{k_i}(x^t_m) d_{\pcd_t} (x^t_m,x^t_n) p_{k_j}(x^t_n)\;,
    \label{eq:pairwise_geodesic}
\end{equation*}
where $d_{\pcd^t}(\cdot,\cdot)$ is the pre-computed geodesic distance between two points in $\pcd^t$. 
If we stack all the keypoint probabilities row-wise in a matrix $W^t \in \mathbb{R}^{K \times N}$ and all the pairwise geodesic distances in a matrix $D^t \in \mathbb{R}^{N\times N}$ with elements $D^t(i,j) = d^t(x^t_i,x^t_j)$, the matrix containing the expected distance between all keypoint pairs can be written as $W^t D^t {W^t}^{\top}$, leading to the geodesic loss:
\begin{equation}
  \mathcal{L}_{geo} = \sum_{\substack{a,b=1 \\ a\neq b}}^{T}\big\| W^a D^{a} {W^a}^{\top} - W^b D^{b} {W^b}^{\top} \big\|_F^2.
 \label{eq:geo_loss} 
\end{equation}
Note that the geodesic distance matrices are needed only during training, and can be created offline during the dataset preprocessing step.

\paragraph{Smoothing loss.} Albeit being a strong prior on the keypoint location, the geodesic distance is not robust to the swap of keypoints due to symmetries on the shape. For instance, swapping the left and right arms of the shape might not bring a significant increase in the geodesic loss.
To alleviate this problem, we propose to pair the geodesic loss with a temporal smoothing loss. 

Under the reasonable assumption of smooth rigid and non-rigid motions, we expect a point of the PCD to not change significantly in its 3D position between two consecutive frames. On the other hand, flipping due to symmetries generally causes an abrupt change in such positions, and thus a larger error for the smoothing loss. We promote the temporal smoothness of the keypoint positions through the following loss:
\begin{equation}
  \mathcal{L}_{smt} = \frac{1}{(T-1)K}  \sum_{t=1}^{T-1} \sum_{j=1}^{K} \big\| \mathbb{E}_{\pcd^{t}}[ k_j] - \mathbb{E}_{\pcd^{t+1}}[k_j] \big\|_2.
 \label{eq:temporal_smoothing} 
\end{equation}

The overall deformation loss is the weighted sum of both its components as described below:
\begin{equation}
  \mathcal{L}_{def} =  \lambda_{geo} \cdot \mathcal{L}_{geo} + \lambda_{smt} \cdot \mathcal{L}_{smt}, 
 \label{eq:temp_loss} 
\end{equation}
where \{$\lambda_{geo}$ and $\lambda_{smt}$\} are hyperparameters experimentally set to $\{6, 2\}$ to equalize the contribution of both losses.

\subsection{Network architecture}
The architecture of the \textit{SelfGeo} is illustrated in  Fig.~\ref{fig:architecture}.
The approach is trained end-to-end. It uses a PointNet++~\cite{ref:qi2017pointnet++} backbone to extract \textit{D}-dimensional features for each point in the input PCD ($\pcd$).
The features are then passed through a Conv1D and a softmax layer to produce the probability matrices $W$ used to compute the geodesic loss.  
The keypoints are further fed to the reconstruction decoder to reconstruct the PCD ($\tilde{\pcd}$). The reconstruction decoder consists of four layers; three Conv1D layers followed by a batch normalization layer and a ReLU activation function, and one Conv1D that produces the reconstructed PCD $\tilde{\pcd}$.

\subsection{Inference and implementation details}
At inference time, the network takes as input just the raw 3D coordinates of a single PCD, without the need to pre-compute any geodesic distances, and it gives as output the expected position of keypoints. We subsampled each PCD to $N=2048$ points and, if not specified differently w.r.t. a category, extract $K=12$ keypoints, 
We implemented \textit{SelfGeo} in PyTorch. 
The models were trained with Adam optimizer, batch size of 32 and learning rate of $1e\text{-}3$ on a 12GB GPU. 

\section{Experiments and evaluation}
\label{sec:experiments_and_evaluation}
This section presents the dataset used in our experiments, performance evaluation metrics, and a comparison of \textit{SelfGeo} with the existing methods.

\subsection{Dataset}
\label{subsec:dataset}
We use three datasets for evaluation:
Clothed Auto Person Encoding (CAPE)~\cite{ref:CAPE:CVPR:20},
Invariant-Top View Dataset (ITOP)~\cite{ref:haque2016towards} and Deforming Things 4D~\cite{ref:li20214dcomplete}. 
The CAPE dataset contains synthetic human models. We considered 61 different temporal sequences divided into train, test, and validation sets as 40 (12432 frames), 11 (3311 frames), and 10 (2228 frames), respectively.
The ITOP dataset contains depth videos captured from the real scene. The videos present human actions. We segment offline the human from the four \textit{side-view} scenes (4626 frames) using the given labels. We observed that the humans are not accurately segmented in several frames, so challenging further the keypoints extraction methods in this real test. 
To validate the claim of generalization, we evaluate \textit{SelfGeo} on the Deforming Things 4D dataset that contains videos of animals performing different actions. We use  9 animals (Bears, Bucks, Bull, Bunny, Chicken, Deer, Dog, Tiger, Rhino) 
for training. We split the animal sequences randomly into 70$\%$, 10$\%$, and 20$\%$, for the training, validation, and testing sets, respectively.

We convert the meshes 
into PCDs of 2048 points normalized within the unit volume.
For training, we precompute geodesic distances for each PCD by creating a graph connecting the five 
neighbours of each point and then approximating the geodesic distance as the shortest path distance between all pairs of points.

\subsection{Metrics for unsupervised keypoints estimation}
\label{subsec:metrics}
We use five metrics to evaluate results on self-supervised keypoints estimation. 
The \textit{coverage} metric evaluates how well the keypoints are distributed over the surface and the \textit{inclusivity} metric indicates how close the keypoints are to the object surface. These metrics have been proposed in previous literature~\cite{ref:fernandez2020unsupervised,ref:zohaib2023sc3k}, and they are a standard measure to evaluate unsupervised keypoints estimation.
The third metric is the \textit{temporal consistency} ($T_{con}$), which shows if keypoints are switching their order in consecutive frames. We compute this metric as: 
\begin{equation}
  \begin{aligned}
   T_{con} &= \frac{100}{(T-1)K} \sum_{t=1}^{T-1} \sum_{i=1}^{K} eq\left(\argmin_j \left(||\mathbb{E}_{\pcd^t} [k_i]- \mathbb{E}_{\pcd^{t+1}}[k_j]||_2 \right), i \right), \\    
  \end{aligned} 
 \label{eq:metric_consistency} 
\end{equation}
where $eq(x,y)$ is 1 if $x=y$ and $0$ otherwise. $T_{con}$ represents the percentage of the semantically consistent keypoints.

The fourth metric is the Probability of Correct Keypoints (PCK), which defines keypoints repeatability across frames. Considering a keypoint $k_i$, we consider its expected position in the first frame $\mathbb{E}_{\pcd^1}[k_i]$ as a reference and compute its repeatability error in a subsequent frame $t$ as: 
\begin{equation}
  \begin{aligned}
  E_{rep}(k_i,t) = \big\| GT_t(\mathbb{E}_{\pcd^1}[k_i]) - \mathbb{E}_{\pcd^t}[k_i]  \big\|_2,
  \end{aligned}
\end{equation} 
where $GT_t(\mathbb{E}_{\pcd^1}[k_i])$ is a function that computes the expected value of keypoint $k_i$ on frame $t$ by transferring its probability distribution over $\pcd^1$ to $\pcd^t$ using the ground-truth point-wise map. The $\textrm{PCK}_{\tau}$ measure is computed as the percentage of keypoints on all subsequent frames $t=2\dots T$  with an error smaller than $\tau$.

Moreover, we consider a 3D reconstruction metric as a downstream task to assess the representation power of the keypoints in encoding the PCD of the deforming shape. For the evaluation of the reconstructed PCD w.r.t. the input PCD, we use the \textit{Chamfer distance} defined in Eq.~\ref{eq:reconstruction}. 

\subsection{Results and analysis}
\label{subsec:results}
We evaluate \textit{SelfGeo} against the SOTA un-/self-supervised keypoints estimation approaches, ULCS~\cite{ref:fernandez2020unsupervised} which estimates 3D keypoints from non-rigid objects under an arbitrary deformation (irrespective of body structure), as we do, and SC3K~\cite{ref:zohaib2023sc3k} which estimates keypoints from an arbitrary posed objects. 
We train all the methods on the CAPE dataset. 
When training of the ULCS and SC3K, we considered the hyperparameter as provided in the respective papers.
For performance evaluation, we test them on synthetic PCDs 
of the CAPE dataset, and on real PCDs extracted from the depth acquisitions of the ITOP dataset. 
As mentioned earlier, ITOP is a particularly challenging dataset, since segmented body PCDs are often distorted and with missing parts. 

To show the ability of \textit{SelfGeo} to handle different 
deformable objects, we compare it on the Deforming Thigs 4D dataset,
which is composed of different animal shapes.
The results reported in Table \ref{tab:comparison_humanCAPE_ITOP} highlight that \textit{SelfGeo} outperforms the baselines for all the metrics.
The keypoints estimated by \textit{SelfGeo} are close to the body surface (high inclusivity), cover the whole body (high coverage),
maintain the semantic order across the frames (high temporal consistency), and provide a reasonable shape reconstruction (lower reconstruction error). 
\begin{table}[t]
\centering
\caption{Comparison of \textit{SelfGeo} with the baseline methods on CAPE, ITOP and Deforming Things 4D datasets. \textit{SelfGeo} outperforms the baselines on all the datasets.}
\begin{tabular}{p{0.13\linewidth}ccccc}
\toprule
Dataset & Approach &  Inclusivity $\uparrow$    & Coverage $\uparrow$       & $T_{con}$ $\uparrow$  & Recon. Err. $\downarrow$   \\
\toprule
\multirow{3}{*}{CAPE~\cite{ref:CAPE:CVPR:20}} & ULCS    & 47.82   & 67.59    & 88.12    & 0.156    \\
                      & SC3K     & 79.32  & 86.46  & 70.72      & 0.025        \\
& SelfGeo      & \textbf{85.87}  & \textbf{91.87} & \textbf{90.43}       & \textbf{0.012}        \\
\midrule
\multirow{3}{*}{ITOP~\cite{ref:haque2016towards}} & ULCS  & 42.08      & 64.28                 & 76.47   & 0.221  \\
 & SC3K   & 76.35   & 83.11 & 42.86      & 0.213        \\
 & SelfGeo    & \textbf{84.23}  & \textbf{91.13} & \textbf{80.64}       & \textbf{0.012}       \\
\midrule
\multirow{3}{0.13\linewidth}{Deforming Things 4D~\cite{ref:li20214dcomplete}} & ULCS    & 	48.06 &  82.91 & 	41.75 & 	0.279  \\
 & SC3K    & 	81.7  & 84.874 & 	77.44 & 	0.153 \\
 & SelfGeo     & \textbf{85.22}	 & \textbf{88.11} & 	 \textbf{80.53} & 	\textbf{0.038}   \\
\bottomrule
\end{tabular}
\label{tab:comparison_humanCAPE_ITOP}
\end{table}

\begin{figure}[t]
     \centering
     \begin{subfigure}[b]{0.33\textwidth}
         \centering
         \includegraphics[width=\textwidth]{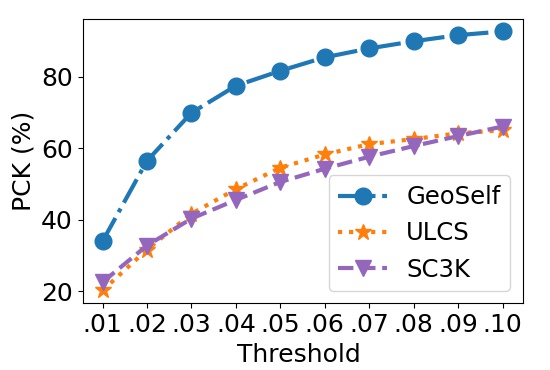}
         \caption{Repeatability test}
         \label{fig:pck_plot}
     \end{subfigure}
     \begin{subfigure}[b]{0.63\textwidth}
         \centering
         \includegraphics[width=\textwidth]{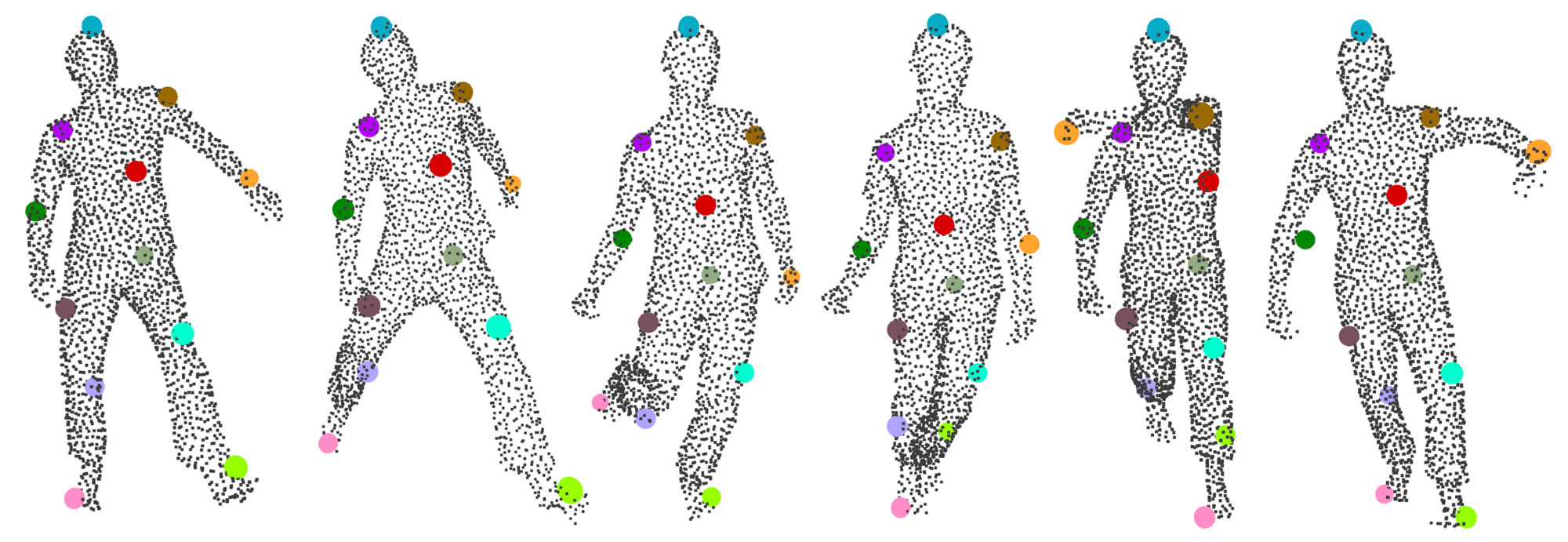}
         \caption{Keypoints estimated by \textit{SelfGeo} on the CAPE dataset}
         \label{fig:comparison_humanCAPE}
     \end{subfigure}
     \caption{The higher PCK in (a) shows that the keypoints estimated by \textit{SelfGeo} have better correspondences across frames, and (b) demonstrates their temporal consistency.
     }
     \label{fig:update_it}
\end{figure}

The repeatability test on the Deforming Things 4D dataset is depicted in Fig.~\ref{fig:pck_plot}, which shows the PCK curves computed by evaluating the $\textrm{PCK}_{\tau}$ measure at increasing threshold distances $\tau$, ranging from 0.01 to 0.10. 
It can be seen that the \textit{SelfGeo} outperforms ULCS and SC3K by a significant margin.

The qualitative results of 
the CAPE dataset are illustrated in Fig.~\ref{fig:comparison_humanCAPE}.
The figure shows the keypoints estimated by the \textit{SelfGeo} for a person playing soccer. 
The colours and positions of the keypoints represent their semantic 
and geometric information. 
It can be seen that the keypoints are stable throughout the shape motion. 
However, in some cases, we observed that due to a significant variation in the human pose, 
the keypoints do not preserve the semantic order. Still, even if the switch happens, a keypoint remains geometrically consistent with a position in the same region of the shape. 
This is likely due to the noise in computing the geodesic distances, which is discussed in supplementary material.

\begin{figure}[t]
  \centering
    \includegraphics[width=.95\linewidth]{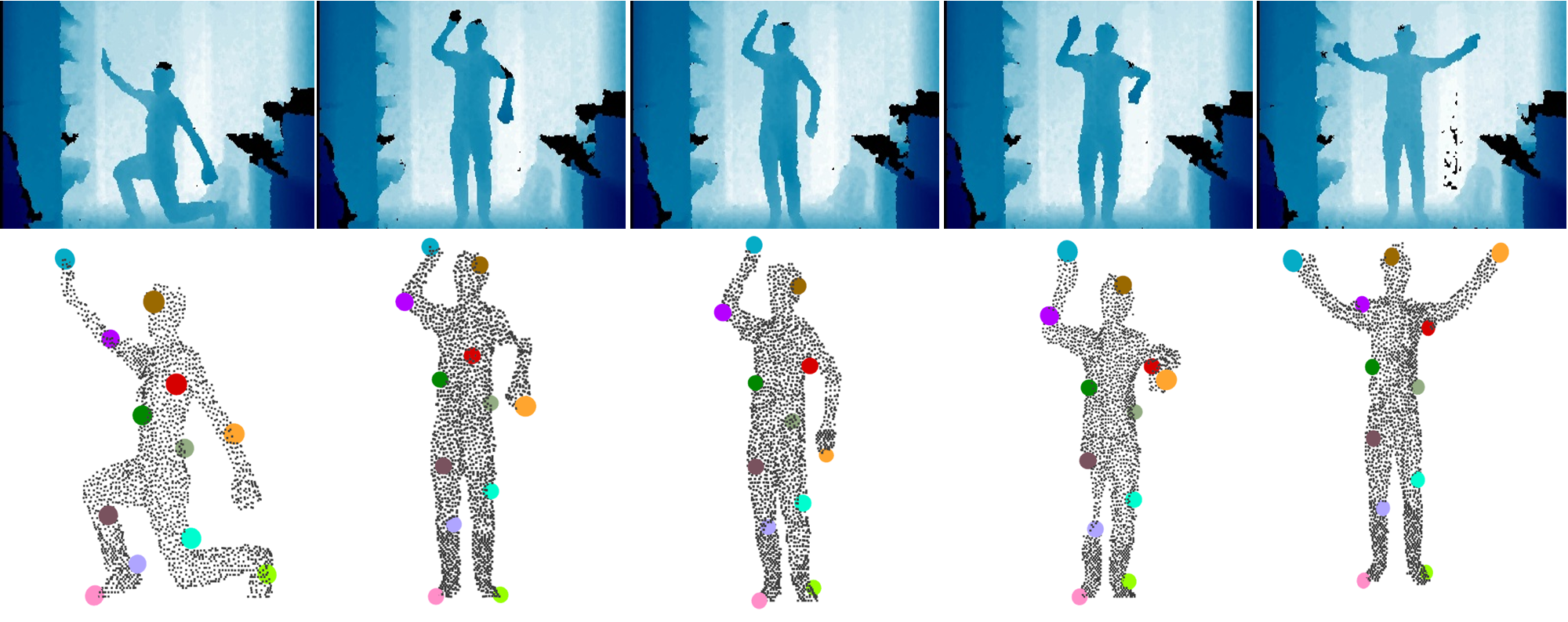}
    \caption{Performance on real humans. The ITOP dataset contains depth images including the background (top row). We segment humans and pass them to the \textit{SelfGeo},~which remains successful in estimating consistent keypoints as shown in the bottom row.}
    \label{fig:comparison_humanITOP}
\end{figure}

The qualitative results for the ITOP datasets are illustrated in Fig.~\ref{fig:comparison_humanITOP}, where the top row shows the original depth images containing a human in a real scene. 
We feed the segmented humans to \textit{SelfGeo} that estimates the keypoints. The estimated keypoints on the input PCD are shown in the bottom row. 
The \textit{SelfGeo} remains successful in estimating accurate keypoints from real sequences. 

\begin{figure*}[ht]
  \centering
   \includegraphics[width=.98\linewidth]{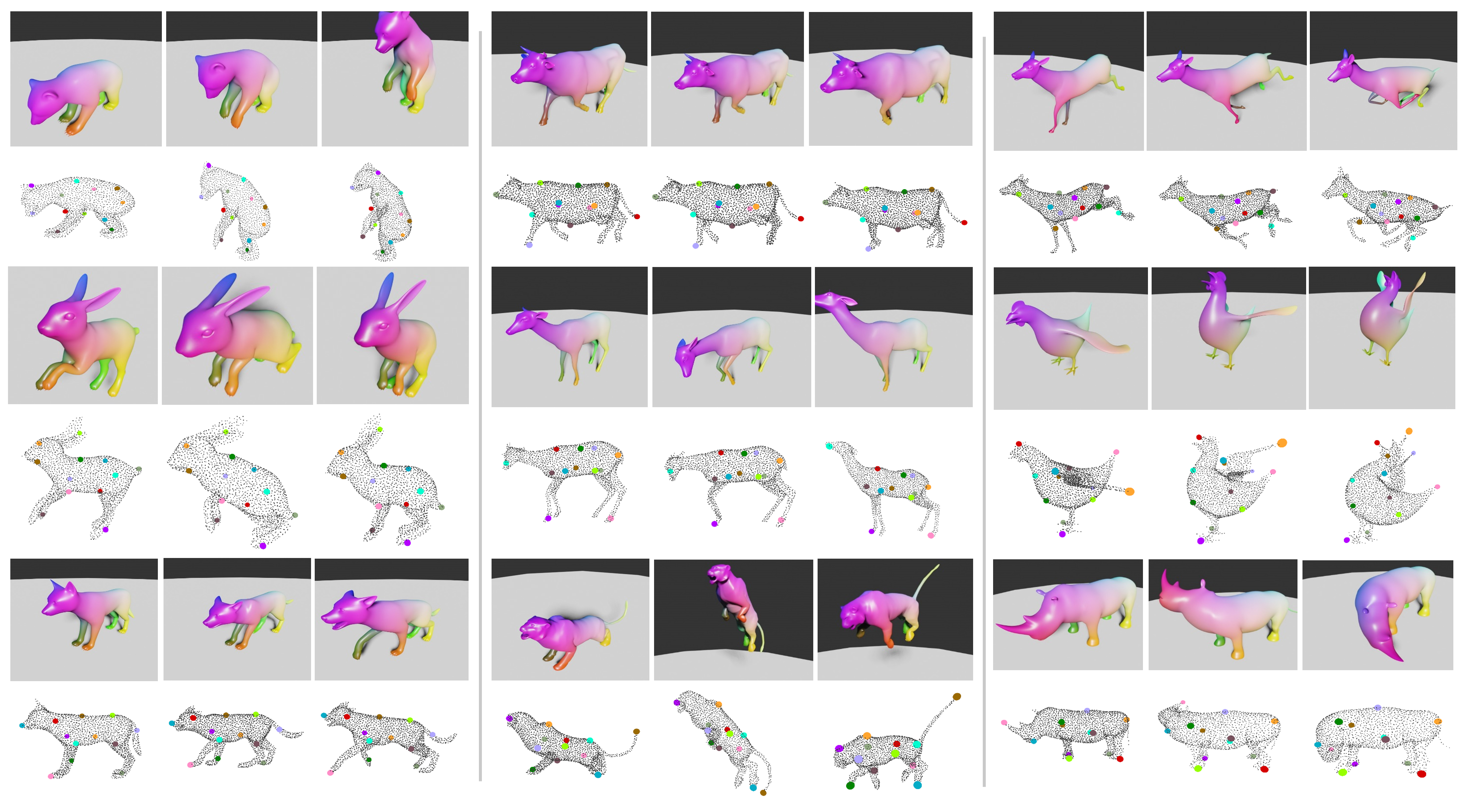}
    \caption{Keypoints estimated on the Deforming Things 4D dataset. 
    The first, third and fifth rows (from left to right) show animals performing different actions. Corresponding estimated keypoints on the input PCDs are illustrated in the second, fourth and sixth rows. The keypoints/PCDs are shown from the side view for a better visualization.}
\label{fig:comparison_deformingThings4D}
\end{figure*}

The qualitative results for the Deforming Things 4D dataset are illustrated in Fig.~\ref{fig:comparison_deformingThings4D}. 
The first, third and fifth rows show
animals performing different actions (jump, run, attack, etc.). The second, fourth and sixth rows illustrate the corresponding estimated keypoints, respectively. The keypoints are temporally consistent even if the range of the motion is quite relevant, denoting that the geodesic loss helps to obtain consistent keypoints that can account for relevant deformations.
Thus the results highlight that, 
on average, \textit{SelfGeo} is robust enough to estimate stable keypoints irrespective of body deformation.
\begin{table}[t]
\centering
\caption{Robustness to the PCD perturbations. 
The \textit{SelfGeo} pretrained on the CAPE dataset is tested for the noisy and downsampled PCDs for different noise variances and sampling ratios. It has a stronger resilience to decimated PCDs and noisy PCDs.}
\begin{adjustbox}{width=.98\columnwidth}
\begin{tabular}{cccccc}
\toprule
 & Inclusivity  $\uparrow$ & Coverage  $\uparrow$ & $T_{con}$  $\uparrow$ & Recon. Err.  $\downarrow$ & GD Err. $\downarrow$\\
\midrule
Variance/Ratio & \multicolumn{5}{c}{Nosiy/Downsampled PCDs}         \\                            \midrule
0.01/x2            & 81.20/82.67        & 90.79/90.85     & 90.23/90.02         & 0.0135/0.0128             & 0.0449/0.012  \\
0.02/x4           & 70.46/78.88        & 89.84/89.33    & 88.08/89.32         & 0.0171/0.0144           & 0.0531/0.025 \\
0.03/x8           & 64.29/75.61        & 88.25/88.72    & 87.74/88.27          & 0.0213/0.0168           & 0.0586/0.047 \\
0.04/x16          & 75.97/73.48        & 85.97/88.19     & 85.67/86.11         & 0.0249/0.0214        & 0.0623/0.069 \\
0.05/x32           & 54.07/62.13        & 82.81/87.85     & 83.93/74.32         & 0.0285/0.0332          & 0.0711/0.089 \\
\bottomrule
\end{tabular}
\end{adjustbox}
\label{tab:noisy_downsampled_test}
\end{table}

To further validate the robustness of the \textit{SelfGeo}, we also evaluate its resilience to noisy and decimated PCDs of the CAPE dataset. To generate the noisy PCDs, we add Gaussian noise with different variances to the original PCDs. For decimating PCDs, we downsampled the original PCDs using the Farthest Point Sampling (FPS) as used in~\cite{ref:mohammadi2021pointview,ref:zohaib2023sc3k}. This experiment simulates the common issues when using different depth/3D sensors with different accuracies and data points densities.   
The results presented in  Tab.~\ref{tab:noisy_downsampled_test} shows that
the \textit{SelfGeo} has a stronger resilience to decimated PCDs than the noisy case, as the x32 downsampling is extremely aggressive. 
This is because the downsampled PCDs maintain the object's structure.
In contrast, introducing noise of a high variance (>0.02) 
changes the shape of the objects, thus reducing the performance.
We observed that when the noise variance is 0.02, the SelfGeo has estimated keypoints that cover the object and hence are good for the reconstruction task (i.e. the error is 0.0144, which is only 0.0016 greater than that when the variance is 0.01).  

\section{Ablations}
\label{sec:ablations}
This section provides two main ablation experiments for \textit{SelfGeo}. First, we train it by excluding one loss at a time to understand the impact of each loss on the approach. Second, we train it for the PointNet~\cite{ref:qi2017pointnet} backbone.
\\
\textbf{Contribution of each loss.}
To test the contribution of each loss, we train \textit{SelfGeo} on the mini-CAPE dataset selected from a part of the CAPE dataset. Mini-CAPE contains 405, 195, and 270 frames in each training, validation and testing set, respectively. At each training round, we ignore one loss
and report results in Tab.~\ref{tab:ablation_losses}. 
We do not show the test without the coverage ($\mathcal{L}_{cov}$) and shape ($\mathcal{L}_{sha}$) loss as their removal makes \textit{SelfGeo} fail, denoting the cardinal importance of these elements.
If we ignore surface loss ($\mathcal{L}_{surf}$), the inclusivity, coverage and $T_{con}$ reduces by 8.66, 6.45 and 5.01, respectively. 
It is due to the fact that they are estimated at random positions, 
mostly outside the shape because of the coverage loss. 
Since they could not preserve the object's geometrical structure, they do not support the 
reconstruction task -- reconstruction error is increased by 0.007. 
Ignoring the reconstruction loss ($\mathcal{L}_{rec}$) decreases the keypoints accuracy in terms of coverage (5.74) and inclusivity (5.91) -- they are estimated in the surroundings of the shape. However, their consistency in the consecutive frames is slightly affected (0.72). This shows that this loss has less influence on the temporal consistency of the keypoints. 
Considering that the smoothing loss ($\mathcal{L}_{smt}$) allows maintaining the distance between the keypoints in the consecutive frames, its ignorance reduces the semantic correspondences between the keypoints ($T_{con}$ is decreased by 2.11). 
In the same way, removing the geodesic distance loss ($\mathcal{L}_{geo}$) reduces also the keypoints localization in different frames (inclusivity and coverage are decreased by 3.74 and 4.14, respectively).  
While removing the deformation loss (
$\mathcal{L}_{smt}$ and $\mathcal{L}_{geo}$) greatly reduces the performance of the network. Not only the inclusivity and coverage are further decreased, but also the $T_{con}$ is decreased with an increase in the reconstruction error.
\begin{table}[t]
\centering
\caption{Ablation on the loss functions (first five rows) and backbone (last row). We train \textit{SelfGeo} by removing one loss 
at each training round
on the mini-CAPE dataset. The increase or decrease in performance is given in the + or - sign, respectively. The colours also code the extent of the performance drop from green to red. The last row highlights that changing the backbone decreases the performance for all the metrics.}
\begin{tabular}{lccccc}
\toprule
Removed Loss & Inclusivity  $\uparrow$ & Coverage    $\uparrow$       & $T_{con} \uparrow$      & Recon.   Err.  $\downarrow$       & GD Err. $\downarrow$                       \\
\midrule
Surface        & \cellcolor[HTML]{F8696B}-8.66 & \cellcolor[HTML]{F8696B}-6.45 & \cellcolor[HTML]{F8696B}-5.01 & \cellcolor[HTML]{FFEA84}+0.007  & \cellcolor[HTML]{69BF7B}+2.8E-05 \\
Reconstruction & \cellcolor[HTML]{FCB579}-5.91 & \cellcolor[HTML]{FFEB84}-5.74 & \cellcolor[HTML]{63BE7B}-0.72 & \cellcolor[HTML]{F8696B}+0.230 & \cellcolor[HTML]{63BE7B}+2.1E-05 \\
Smoothing      & \cellcolor[HTML]{63BE7B}-3.36 & \cellcolor[HTML]{63BE7B}-2.72 & \cellcolor[HTML]{B7D780}-2.11 & \cellcolor[HTML]{63BE7B}+0.001 &  \cellcolor[HTML]{FFEB84}+1.8E-04 \\
Geodesic       & \cellcolor[HTML]{C2DA81}-3.74 & \cellcolor[HTML]{ADD480}-4.14 & \cellcolor[HTML]{FFEB84}-3.31 & \cellcolor[HTML]{97CD7E}+0.002 &  \cellcolor[HTML]{FFEA84}+2.6E-03 \\
Deformation   & \cellcolor[HTML]{FFEB84}-3.99 & \cellcolor[HTML]{FDC67C}-5.94 & \cellcolor[HTML]{FCC37C}-3.82 & \cellcolor[HTML]{FFEB84}+0.004 & \cellcolor[HTML]{F8696B}+2.1E-01 \\
\midrule
PointNet backbone   & -1.31  & -0.53	 & -1.95 & 	+0.002 & +1.75E-03 \\
\bottomrule
\end{tabular}
\label{tab:ablation_losses}
\end{table}
\\
\textbf{Impact of the backbone.}
We provide a further ablation for two widely used backbones; PointNet and PointNet++. We observed that the performance of the \textit{SelfGeo} is higher when PointNet++ backbone is used. The last row of Tab.~\ref{tab:ablation_losses} depicts the performance drop when PointNet encoder is integrated in \textit{SelfGeo}. 

\section{Conclusions}
\label{sec:conclusions}
In this paper, we introduce \textit{SelfGeo}, a novel method to estimate the 3D keypoints on the deformable shapes in a self-supervised way
without requiring the ground truth labels. 
Such keypoints should remain temporally consistent across the 3D frames, irrespective of the shape deformation. \textit{SelfGeo} achieved this goal by considering the invariant properties of the deforming shapes -- the keypoints should deform along with the non-rigid motion of the shape (semantically consistent), however, they should also maintain a constant geodesic distance among them (geometrically consistent). 
We evaluate our method against the SOTA approaches on three different deformable datasets. 
The results demonstrate the superiority of \textit{GeoSlef} over the baselines. Morevoer, we also present the test on the noisy and decimated PCDs. It shows that the performance of the \textit{SelfGeo} has slightly decreased; however, it remains successful in estimating the consistent keypoints.
\\
\textbf{Limitations of the proposed approach.}
Noise in the geodesics computation can affect our method. We have observed that when two body parts touch each other so that the 3D points are so close, they are considered connected (and they should not be). 
Thus, the geodesic distance between them approaches zero, which is an error. 
When this error happens, the keypoints estimated by \textit{SelfGeo} might change their positions.
See a practical example in the supplementary material. 
Moreover, symmetry in the object's parts, which is considered a basic problem in 3D vision, also reduces the network's performance. 
\\
\vspace{0.5cm}
\\
\textbf{Acknowledgements:}
We would like to acknowledge Pietro Morerio for fruitful discussions. 
This work was carried out within the frameworks of the project ``RAISE - Robotics, and AI for Socio-economic Empowerment'' and the PRIN 2022 project n. 2022AL45R2 (EYE-FI.AI, CUP H53D2300350-0001). This work has been supported by European Union - NextGenerationEU.


%
%
\bibliographystyle{splncs04}
\bibliography{main}

%
%
%
%
%
%
%
%
%
%
%
%
%
%
%
%
%
%
%
%
%
%
%
%
%
%
%
%
%
%
%
%
%
%
%
%
%
%
%
%
%
%
%
%
%
%
%
%
%
%
%
%
%
%
%
%
%
%
%
%
%
%
%
%
%
%
%
%
%
%
%
%
%
%
%
%
%
%
%
%
%
%
%
%
%
%
%
%
%
%
%
%
%
%
%
%
%
%
%
%
%
%
%
%
%
%
%
%
\clearpage
\appendix
\begin{minipage}{0.9\columnwidth}
\centering{\textbf{\LARGE Supplementary Material}}
\vspace{1cm}
\end{minipage}
\vspace{0.3cm}
\\
Here we present supplementary material for the ``SelfGeo: 
Self-supervised and Geodesic-consistent Estimation of Keypoints on Deformable Shapes''. 
It presents the qualitative comparison of \textit{SelfGeo} and the State-of-the-art (SOTA) methods based on the CAPE dataset, quantitivate results on the Dynamic FAUST dataset, selection criteria of the total number of estimated keypoints, and highlights limitations of using geodesic distances that prevent \textit{SelfGeo} in estimating stable keypoints in the presence of topological deformations. 
Furthermore, it also presents additional ablations based on the data augmentation, provides further qualitative visualizations for the easy understanding of the robustness of the \textit{SelfGeo}.

\section{Qualitative comparison with the SOTA methods:} 
\begin{figure}[h]
  \centering
    \includegraphics[width=.70\linewidth]{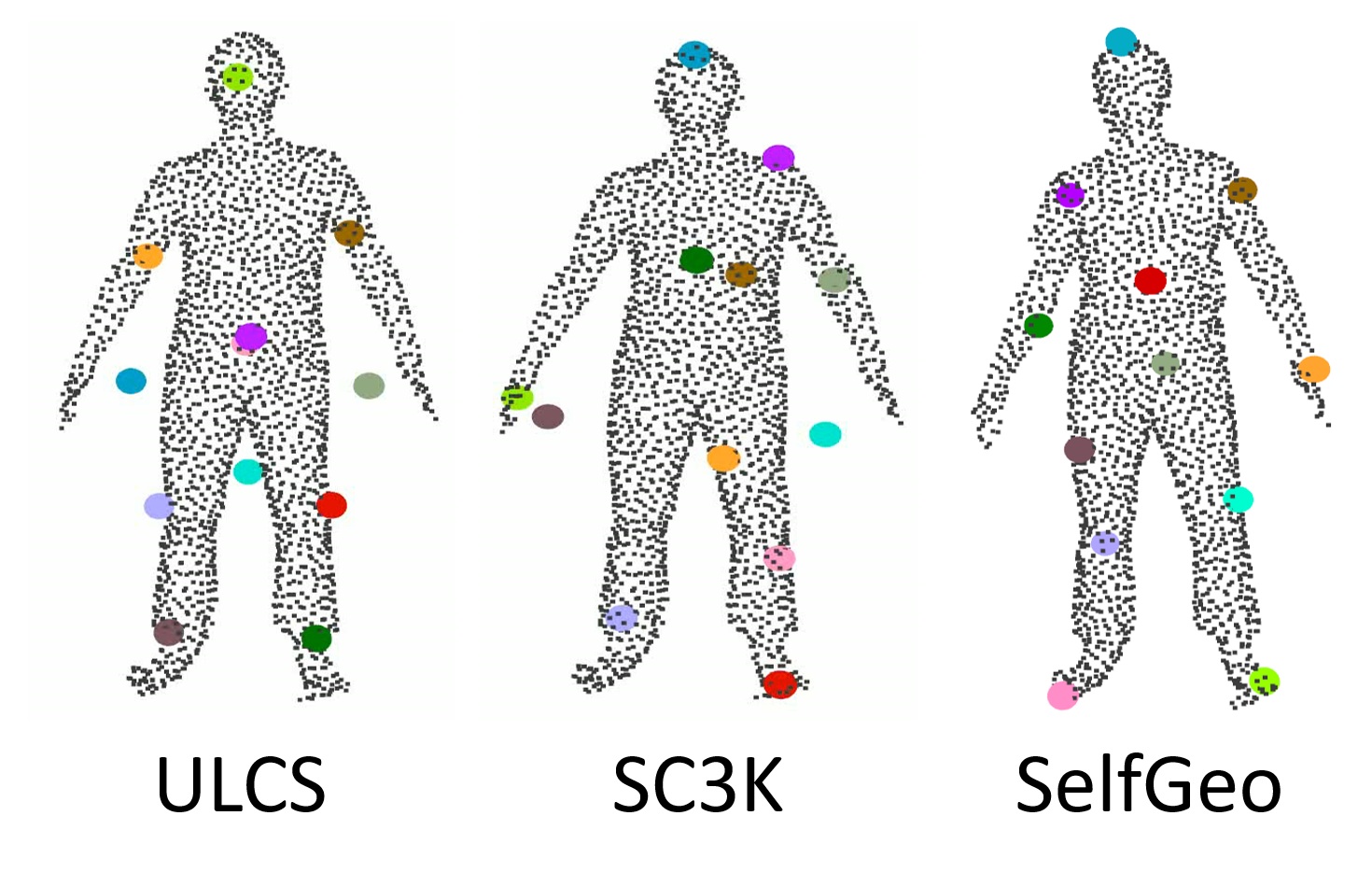}
  \caption{Qualitative comparison with the SOTA methods}
  \label{fig:comparison_baseline}
\end{figure}

A qualitative comparison is shown in Fig.~\ref{fig:comparison_baseline}. We observed that the keypoints estimated by the SOTA methods are not semantically or geometrically consistent for deformable shapes/sequences and often lie outside of the surface. 
%
%
%
%
%
%
%
%
%

\section{Comparison on Dynamic FAUST dataset:} 
We used CAPE dataset in our experiments since it contains 4D sequences with a high level of deformation in comparison with the Dynamic FAUST.
Nevertheless, Tab.~\ref{tab:reb_comparison_dfaust} shows the performance of models trained on CAPE when tested on Dynamic FAUST, demonstrating similar results and thus further validating the generalizability of our method.
Note that the results reported in the ULCS paper for Dynamic FAUST are trained and tested on the same sequence (action), so test shapes are very similar to the ones seen during training. 
%
%
%
\begin{table}[t]
\centering
\caption{Comparison on Dynamic FAUST dataset.} 
%
\begin{tabular}{p{0.25\linewidth}cccc}
\toprule
Approach &  Inclusivity $\uparrow$    & Coverage $\uparrow$       & $T_{con}$ $\uparrow$  & Recon. Err. $\downarrow$   \\
\midrule
ULCS     & 68.15 & 73.96	 & 	73.28 & 0.034
\\
SC3K    & 	83.33 & 85.63 & 	82.72 & 	0.031     \\
\textit{SelfGeo}  & 	\textbf{90.72}	 & \textbf{91.23} & \textbf{93.76} & 	\textbf{0.019} \\
\bottomrule
\end{tabular}
\label{tab:reb_comparison_dfaust}
\end{table}
%
%
%
%
%
%
\begin{figure}[t]
  \centering
  \includegraphics[width=.95\linewidth]{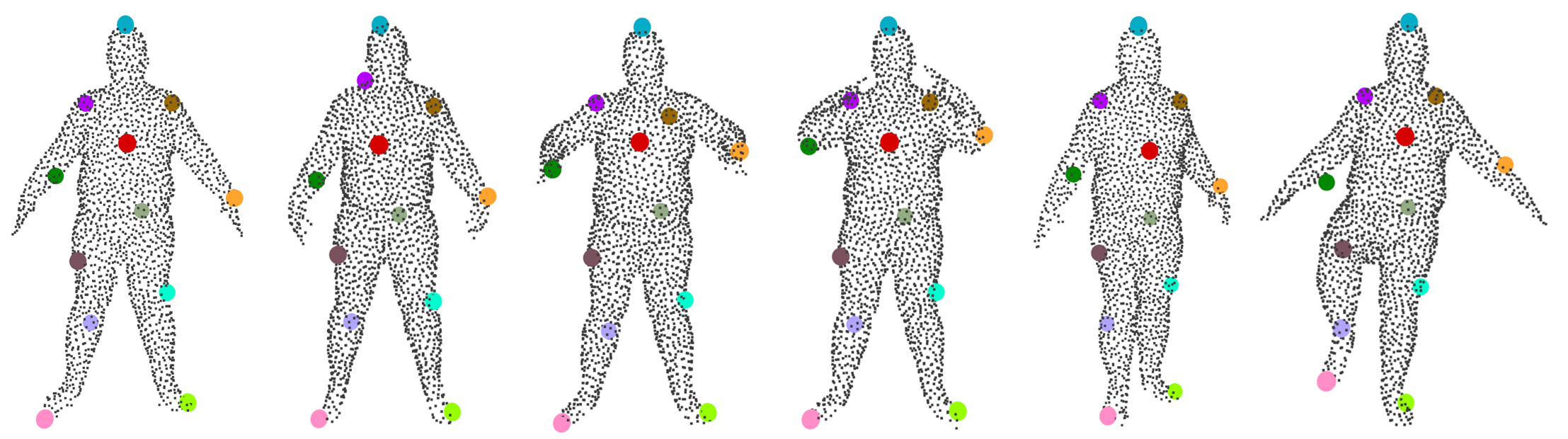}
  \caption{Keypoints estimated by the \textit{SelfGeo} on the DFAUST dataset.}
  \label{fig:comparison_dfaust}
\end{figure}
%
%
Qualitative results of the \textit{SelfGeo} on the DFAUST dataset are illustrated in Fig.~\ref{fig:comparison_dfaust}.

\section{Selection of the total number of keypoints:} 
\begin{figure}[h]
  \centering
    \includegraphics[width=.90\linewidth]{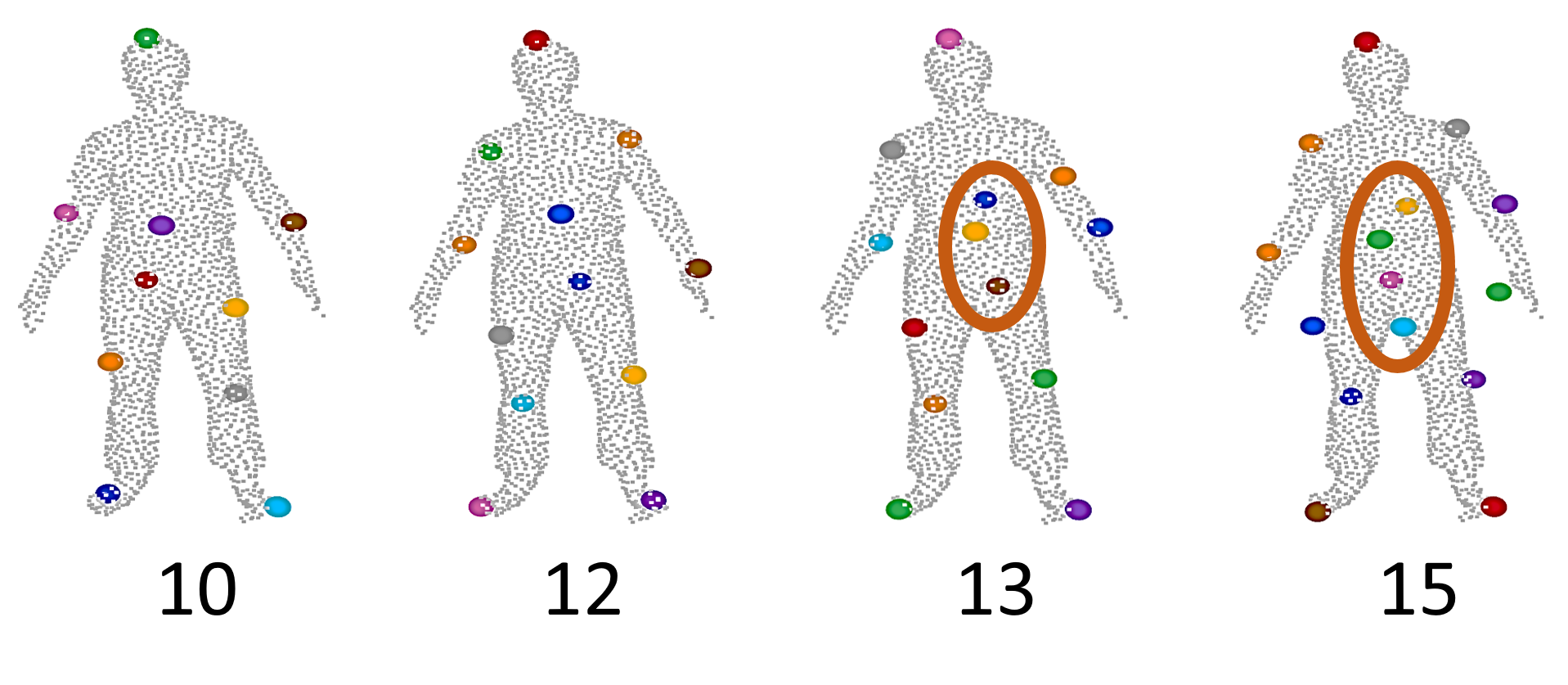}
  \caption{\textit{SelfGeo} for different number of keypoints.}
  \label{fig:diff_kpts}
\end{figure}

We experimented with different numbers of keypoints and observed that 12 keypoints best describe the object's shape. The qualitative results are illustrated in Fig.~\ref{fig:diff_kpts}.
It is observed that fewer keypoints miss the important parts of the objects (e.g. shoulders), and more than 12 keypoints are estimated close to each other,
as also reported in SC3K~\cite{ref:zohaib2023sc3k}.
%
%
%
%
%

%
%
%
%

\section{Geodesic distances and their limitations}
\begin{figure}[!h]
  \centering
    \includegraphics[width=.90\linewidth]{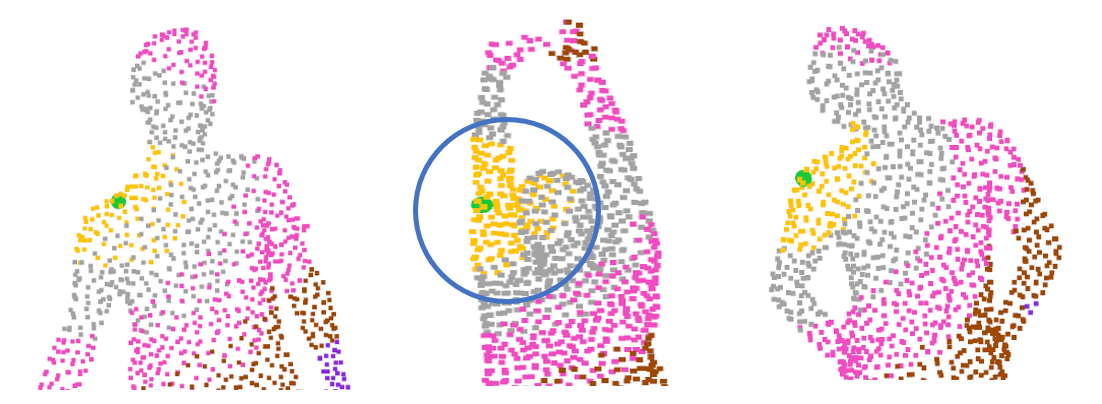}
  \caption{Error in geodesic distances. The green point is a reference point (for the three objects) from where the geodesic distances are computed. The yellow small points lie within the same distance limit w.r.t. the reference point. Ideally, these yellow points should highlight the same region on all the objects irrespective of deformation. It can be viewed that in the left or in the right object, these yellow points are on the left shoulder.
  However, in the center object (inside the blue circle),
  some of these yellow points are on the head, which is an error.
  }
  \label{fig:gd_error}
\end{figure}

Although geodesic distance gives a strong clue about the body deformation, it has some limitations.
The primary issue arises when handling deformations that cause the object's surface to be very near or intersect in points that are contiguous in the object. 
In such scenarios, the geodesic path constructed on the PCD will jump from one point on the surface to the other, resulting in the computation of a wrong geodesic distance. 
We show a practical example in Fig.~\ref{fig:gd_error} on how this can affect the geodesic distance computation, resulting in noise in the dataset that might affect the keypoint localization.
In the figure, the geodesic distances from the green point are computed and color-coded as isocurves.  
%
%
%
%
%
%
%
%
%

%
%
%
%

\section{Additional Ablations}
We 
evaluate the robustness of \textit{SelfGeo} to different sources of perturbations on the PCD, such that vertex coordinates perturbation and subsampling. We train the proposed network on the perturbed PCDs and then test it on both clean and perturbed PCDs on the mini-CAPE dataset. Results are reported in tables \ref{tab:noisy_pcds_test} and \ref{tab:downsampled_pcds}, indicating the variation with respect to the model trained and tested on clean PCDs.

\subsection{Evaluation on noisy PCDs}
We generate noisy PCDs by adding isotropic Gaussian noise with different variances to the vertex coordinates of the original clean PCDs. After training \textit{SelfGeo} on the noisy version, we test the model on both clean and noisy test sets. The results (performance drop w.r.t. the results of the network trained without augmentation) are reported in Tab.~\ref{tab:noisy_pcds_test}.
\begin{table}[t]
\centering
\caption{Evaluation on noisy PCDs. The performance is tested on the original (clean) as well as on the perturbed (noisy) test set.
The results show that the performance decreases with the increase in the noise variances.
Moreover, the keypoints estimated on the noisy PCDs are close to objects (high inclusivity), temporally consistent (high $T_{con}$) and maintain the geodesic distances error (low GD Err.).
However, keypoints estimated on the clean PCDs cover more area of the object and are good for the reconstruction task.
}
\begin{tabular}{cccccc}
\toprule
 & Inclusivity  $\uparrow$ & Coverage $\uparrow$ & $T_{con}$ $\uparrow$ & Recons. Err.  $\downarrow$ & GD Err.  $\downarrow$\\
\midrule
Variance & \multicolumn{5}{c}{\textbf{Test for clean PCDs}}         \\                            \midrule
0.02 & -0.61  & -3.75 & -0.03 & 0.004 & 2.46E-04 \\
0.04 & -1.63  & -4.38 & -1.65 & 0.011 & 4.40E-04 \\
0.06 & -17.58 & -4.61 & -1.76 & 0.018 & 2.28E-03 \\
0.08 & -23.97 & -4.72 & -1.88 & 0.029 & 2.88E-03 \\
0.1  & -38.82 & -9.00 & -3.37 & 0.043 & 3.18E-03 \\
\midrule
Variance & \multicolumn{5}{c}{\textbf{Test for noisy PCDs}}         \\                            \midrule
0.02 & -1.31 & -6.62  & -0.07 & 0.0041 & 3.02E-04 \\
0.04 & -1.42 & -11.53 & -0.18 & 0.0071 & 3.61E-04 \\
0.06 & -2.94 & -16.44 & -0.31 & 0.0098 & 3.66E-04 \\
0.08 & -6.40 & -22.24 & -1.33 & 0.0126 & 3.98E-04 \\
0.1  & -8.87 & -27.76 & -1.33 & 0.0153 & 4.52E-04 \\
\bottomrule
\end{tabular}
\label{tab:noisy_pcds_test}
\end{table}

It can be observed that the performance generally decreases with the increase in the noise variance. 
The estimated keypoints are closer to the object's surface when noisy PCDs are used in the test since the keypoints estimated away from the original PCDs will remain close to the noisy PCDs. 
Moreover, the keypoints on the noisy PCDs are temporally more consistent and maintain the geodesic distances throughout the frames.
On the other hand, the keypoints on the clean PCDs cover more area of the object than those estimated on the noisy PCDs, and therefore, they are good candidates for the reconstruction task (lower reconstruction error).

\subsection{Evaluation on downsampled PCDs}
We generate the downsampled versions of the original PCDs using the Farthest Point Sampling (FPS) as described in~\cite{ref:mohammadi2021pointview,ref:zohaib2023sc3k}. We use it to train \textit{SelfGeo} and then test on the original and the downsampled PCDs. The comparison is shown in Tab.~\ref{tab:downsampled_pcds}.
\begin{table}[t]
\centering
\caption{Training on the downsampled PCDs. The \textit{SelfGeo} trained on the downsampled PCDs is tested on original (clean) and downsampled PCDs. 
Overall, the keypoints estimated on the downsampled test set are close to the surface, covering the object, and are good for the reconstruction task. On the other hand, the keypoints estimated on the clean test set are relatively more consistent with lower geodesic distance error.}
\begin{tabular}{cccccc}
\toprule
 & Inclusivity  $\uparrow$ & Coverage $\uparrow$ & $T_{con}$ $\uparrow$ & Recons. Err.  $\downarrow$ & GD Err.  $\downarrow$\\
\midrule
Ratio & \multicolumn{5}{c}{\textbf{Test for clean PCDs}}         \\                            \midrule
x2  & -0.78   & -3.81  & -0.16 & 0.003 & 6.33E-04 \\
x4  & -1.41  & -6.08  & -0.46 & 0.012 & 7.03E-04 \\
x8  & -16.84 & -10.10 & -1.35 & 0.015 & 7.19E-04 \\
x16 & -17.41 & -10.43 & -5.42 & 0.021 & 7.69E-04 \\
x32 & -34.00 & -14.76 & -7.16 & 0.030 & 7.77E-04  \\
\midrule
Ratio & \multicolumn{5}{c}{\textbf{Test for downsampled PCDs}}         \\                            \midrule
x2  & -0.74  & -3.52  & -0.84  & 0.0003 & 7.08E-04 \\
x4  & -1.45  & -5.96  & -2.51  & 0.0038 & 9.00E-04 \\
x8  & -4.32  & -9.77  & -10.28 & 0.0080 & 1.39E-03 \\
x16 & -7.22  & -9.79  & -28.07 & 0.0097 & 1.94E-03 \\
x32 & -25.56 & -11.84 & -40.03 & 0.0138 & 2.61E-03  \\
\bottomrule
\end{tabular}
\label{tab:downsampled_pcds}
\end{table}

The table shows that the keypoints estimated on the downsampled PCD are covering the object, close to the surface and are good for the reconstruction task. The reason is the same as the previous, the network has seen only the augmented PCDs during the training, and hence, it is performing well for them. On the other hand, the keypoints are more temporally consistent with the lower geodesic error when the clean PCDs are used for testing. This is due to the fact that for the dance (clean) PCDs, the convex combination gives more stable keypoints than the sparse PCDs. 

\subsection{Evaluation on the noisy and downsampled PCDs}
In this ablation, we train the \textit{SelfGeo} for the noisy and downsampled PCDs. We fix the downsampling ratio to x2 (points/2) and increase the noise variance for each training. The selection of the downsampling ratio is based on the fact that the performance of the \textit{SelfGeo} drops with the increase in the downsampling ratio, which is found in the previous ablations.
We follow the same convention as we follow in the previous two ablations and test the trained networks for clean and augmented PCDs. The results are provided in the Tab.~\ref{tab:noisy_doensampled_pcds_test}.
\begin{table}[t]
\centering
\caption{Training on the downsampled and noisy PCDs.
We train the \textit{SelfGeo} for downsampled (x2) and noisy PCDs.
The performance is tested on the clean and the augmented test set.
The network trained for x2+0.02 augmentation estimates keypoints that lie on the surface and cover the object, when tested on the clean PCDs. The results are better than the one when no augmentation is used. 
however, they are not comparatively more consistent in consecutive frames.
}
\begin{tabular}{cccccc}
\toprule
 & Inclusivity  $\uparrow$ & Coverage $\uparrow$ & $T_{con}$ $\uparrow$ & Recons. Err.  $\downarrow$ & GD Err.  $\downarrow$\\
\midrule
Variance + Ratio  & \multicolumn{5}{c}{\textbf{Test for clean PCDs}}         \\                            \midrule
x2 + 0.02 & \textbf{0.17} & \textbf{0.21} & -2.25  & 0.0039 & 2.86E-03 \\
x2 + 0.04 & -3.40         & -1.10         & -5.84  & 0.0098 & 4.40E-03 \\
x2 + 0.06 & -3.83         & -3.92         & -7.12  & 0.0138 & 4.54E-03 \\
x2 + 0.08 & -3.96         & -4.57         & -7.14  & 0.0200 & 5.26E-03 \\
x2 + 0.10  & -4.51        & -6.54         & -10.09 & 0.0268 & 5.43E-03  \\
\midrule
Variance + Ratio  & \multicolumn{5}{c}{\textbf{Test for noisy and downsampled PCDs}}         \\       \midrule
x2 + 0.02 & -0.88  & -2.30  & -0.99 & 0.0056 & 2.99E-03 \\
x2 + 0.04 & -5.22  & -5.74  & -2.00 & 0.0087 & 4.21E-03 \\
x2 + 0.06 & -12.91 & -11.93 & -2.60 & 0.0111 & 4.58E-03 \\
x2 + 0.08 & -25.96 & -15.99 & -3.50 & 0.0132 & 4.81E-03 \\
x2 + 0.10  & -39.26 & -18.11 & -4.41 & 0.0156 & 4.91E-03  \\
\bottomrule
\end{tabular}
\label{tab:noisy_doensampled_pcds_test}
\end{table}

This ablation shows that we can estimate the keypoints (on the original/clean PCD) close to the surface that also covers the whole object by adding noise of 0.02 variance to the train set and downsampling it by x2. 
Although the performance of the \textit{SelfGeo} when trained with this augmentation is higher than when it is trained without any augmentation on the scale of inclusivity (+0.17) and coverage (+0.21), 
the keypoints are not temporally consistent and the geodesic distance error is also increased. 
On average, we get favourable keypoints w.r.t. the inclusivity, coverage, and reconstruction error metrics when we test the network on the clean PCDs. And, $T_{con}$ and geodesic distance error is lower when we use the augmented PCDs.

\section{Qualitative results for the Deforming Things 4D dataset}
\begin{figure}[]
  \centering
    \includegraphics[width=.95\linewidth]{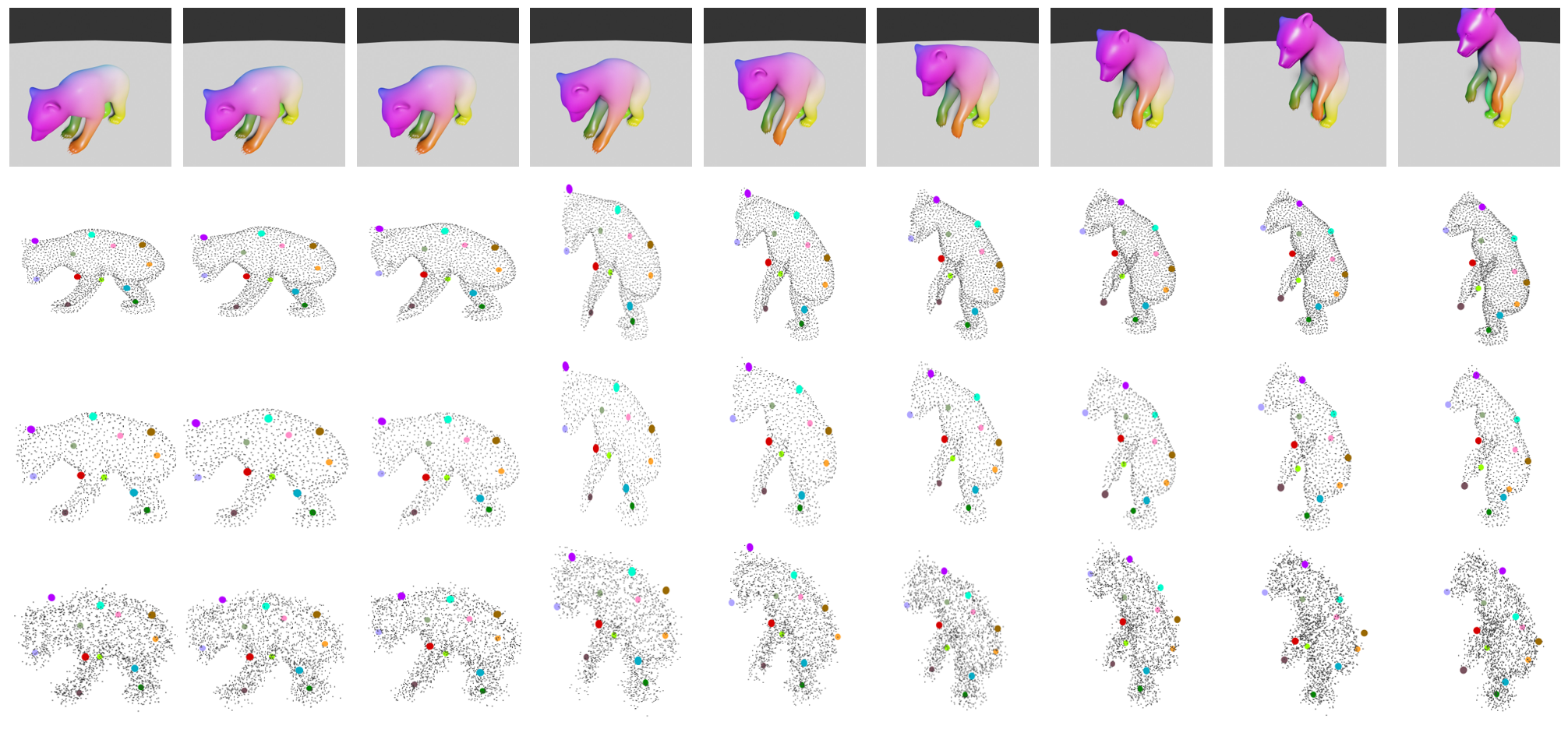}
  \caption{Keypoints estimated by \textit{SelfGeo} on bears.  }
  \label{fig:supp_bear}
\end{figure}

\begin{figure}[]
  \centering
    \includegraphics[width=.95\linewidth]{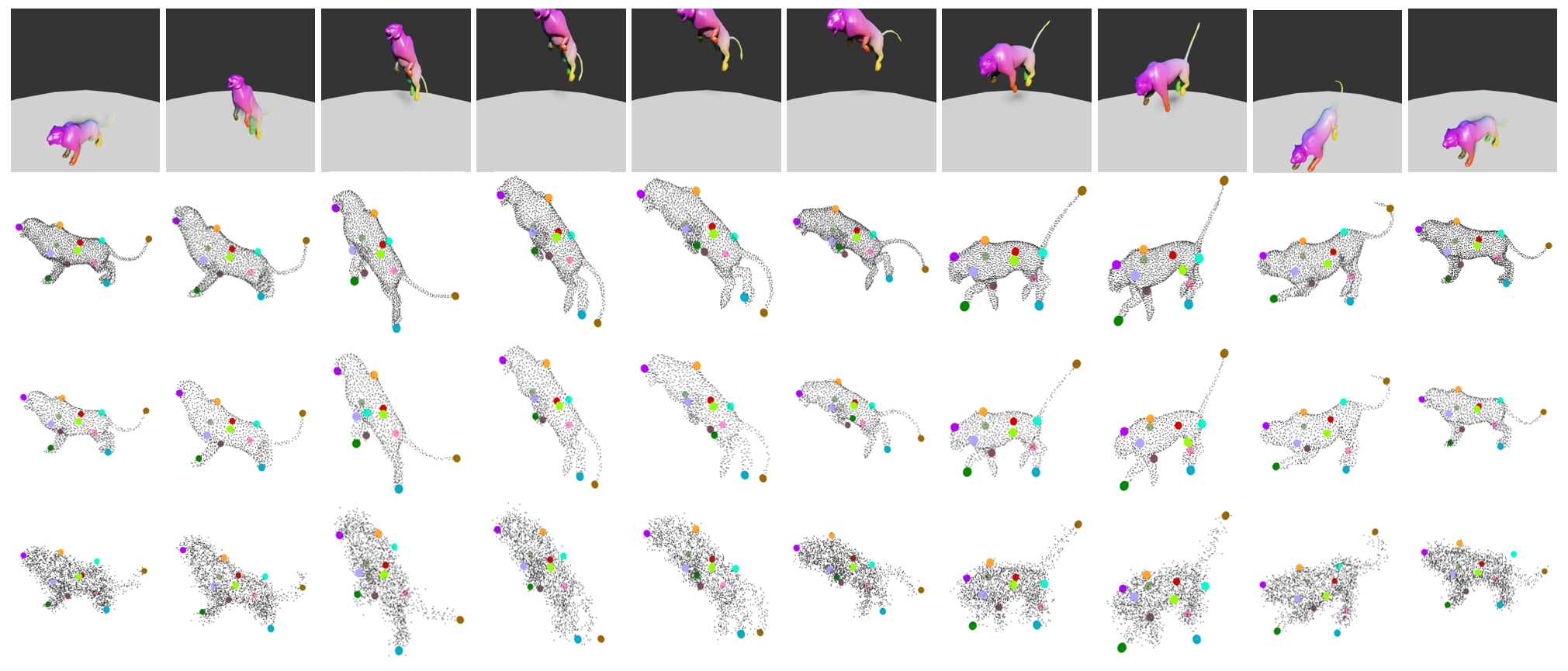}
  \caption{Keypoints estimated by \textit{SelfGeo} on tiger.  }
  \label{fig:supp_tiger}
\end{figure}

\begin{figure}[]
  \centering
    \includegraphics[width=.95\linewidth]{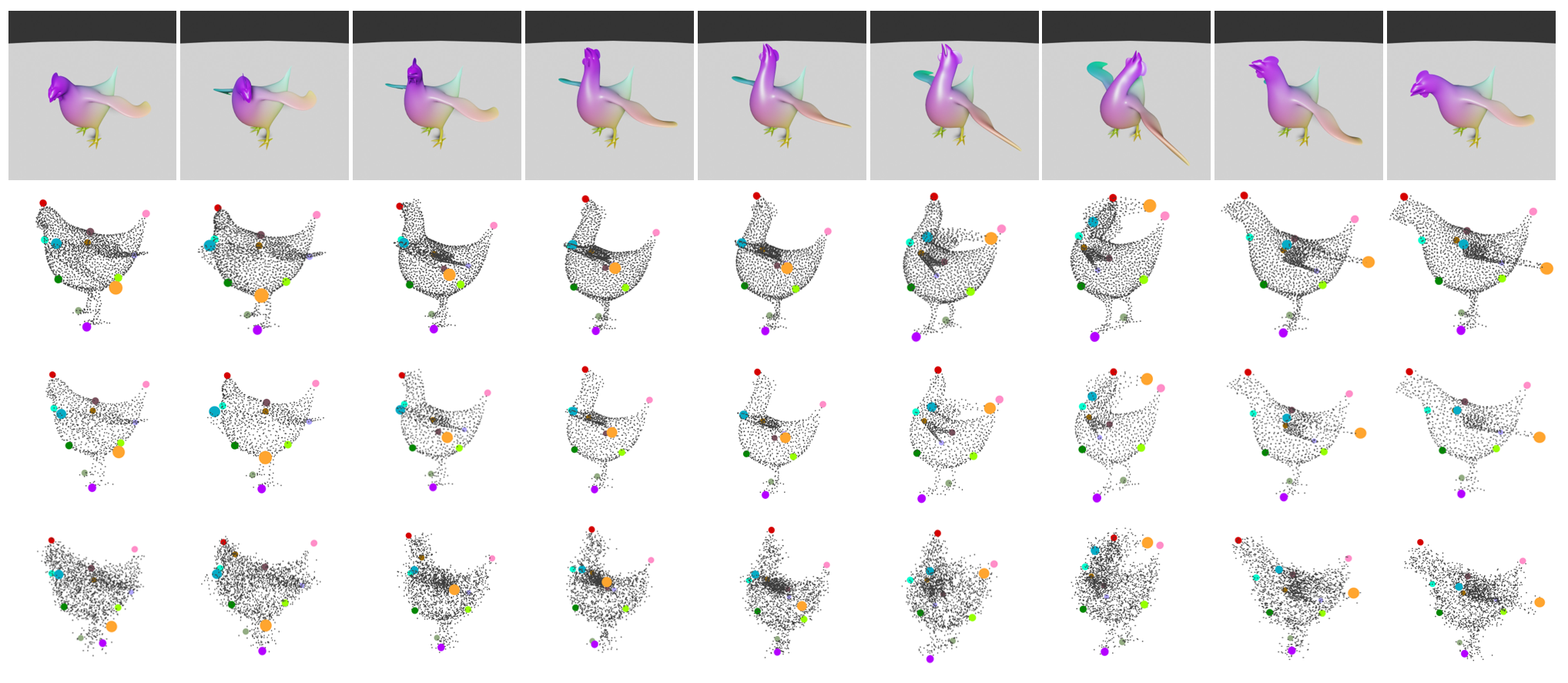}
  \caption{Keypoints estimated by \textit{SelfGeo} on chicken.  }
  \label{fig:supp_chicken}
\end{figure}

\begin{figure}[]
  \centering
    \includegraphics[width=.95\linewidth]{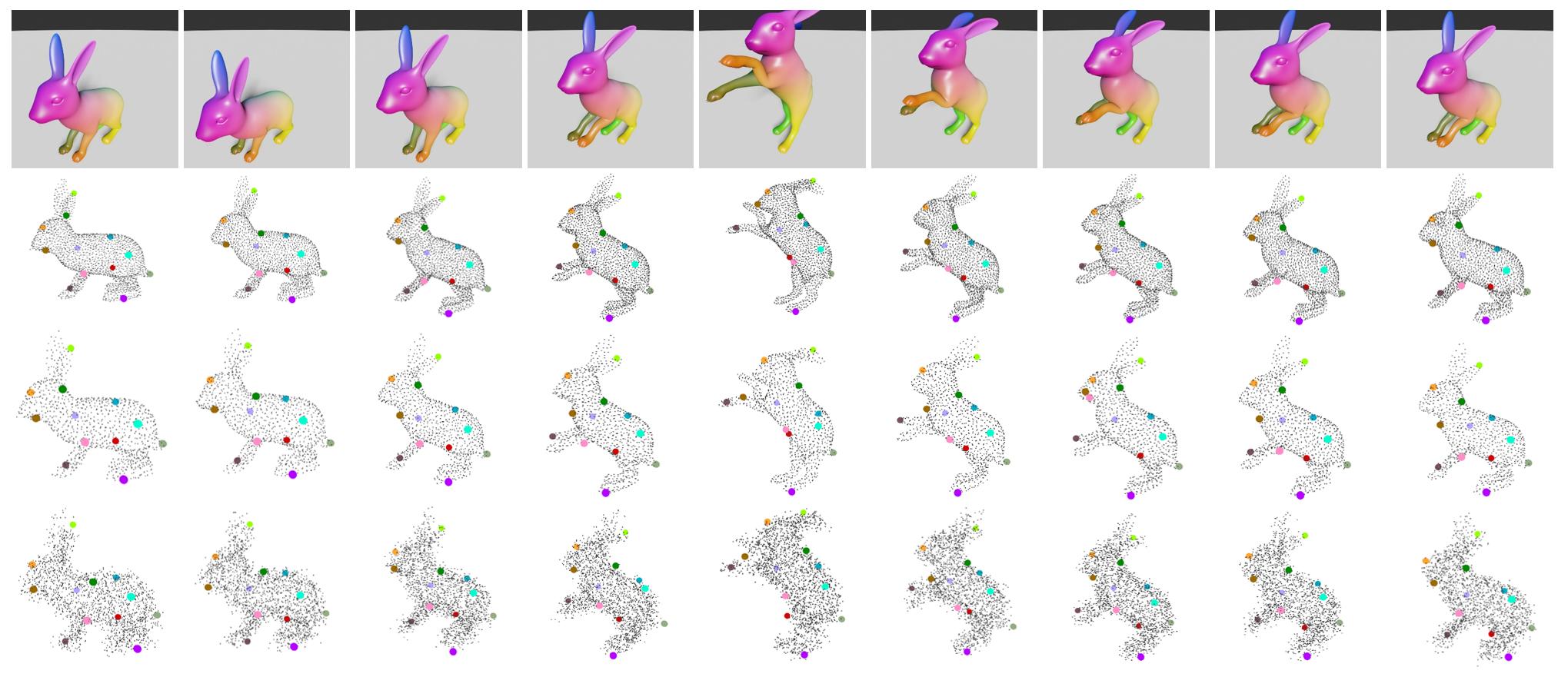}
  \caption{Keypoints estimated by \textit{SelfGeo} on rabbit.  }
  \label{fig:supp_bunny}
\end{figure}

\begin{figure}[]
  \centering
    \includegraphics[width=.95\linewidth]{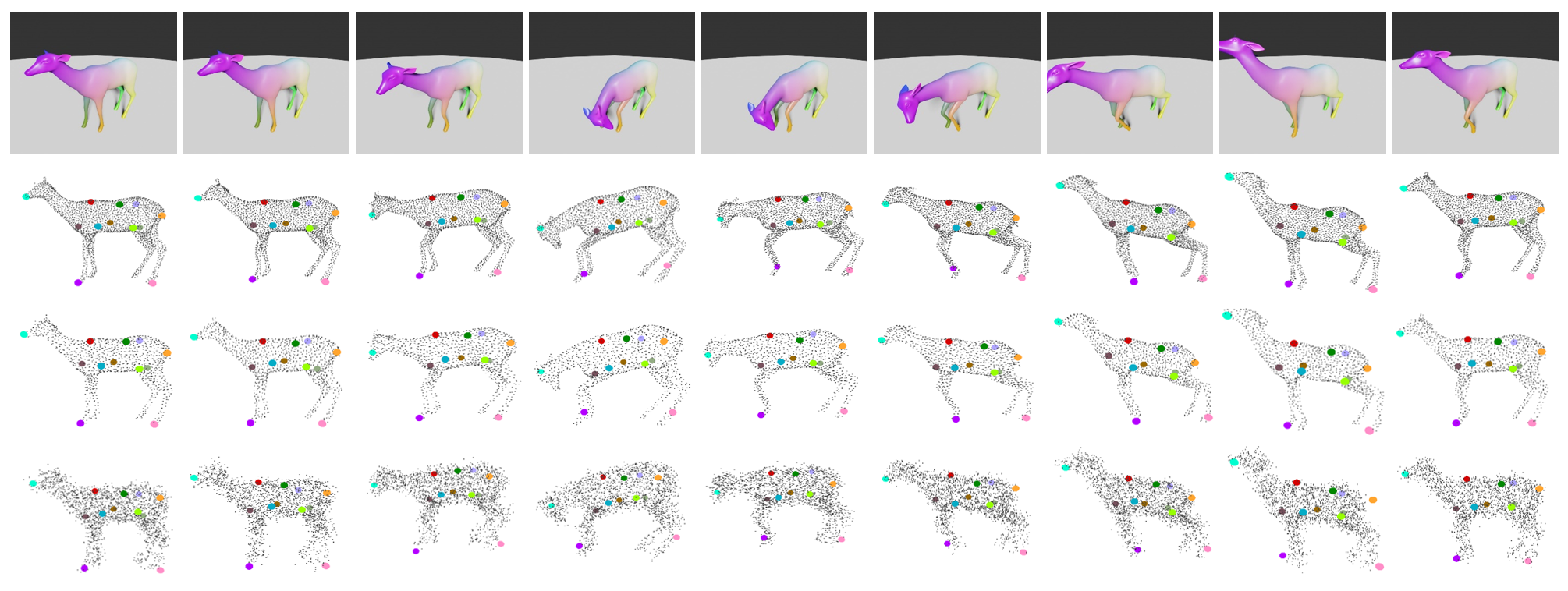}
  \caption{Keypoints estimated by \textit{SelfGeo} on deer.  }
  \label{fig:supp_deer}
\end{figure}

\begin{figure}[]
  \centering
    \includegraphics[width=.95\linewidth]{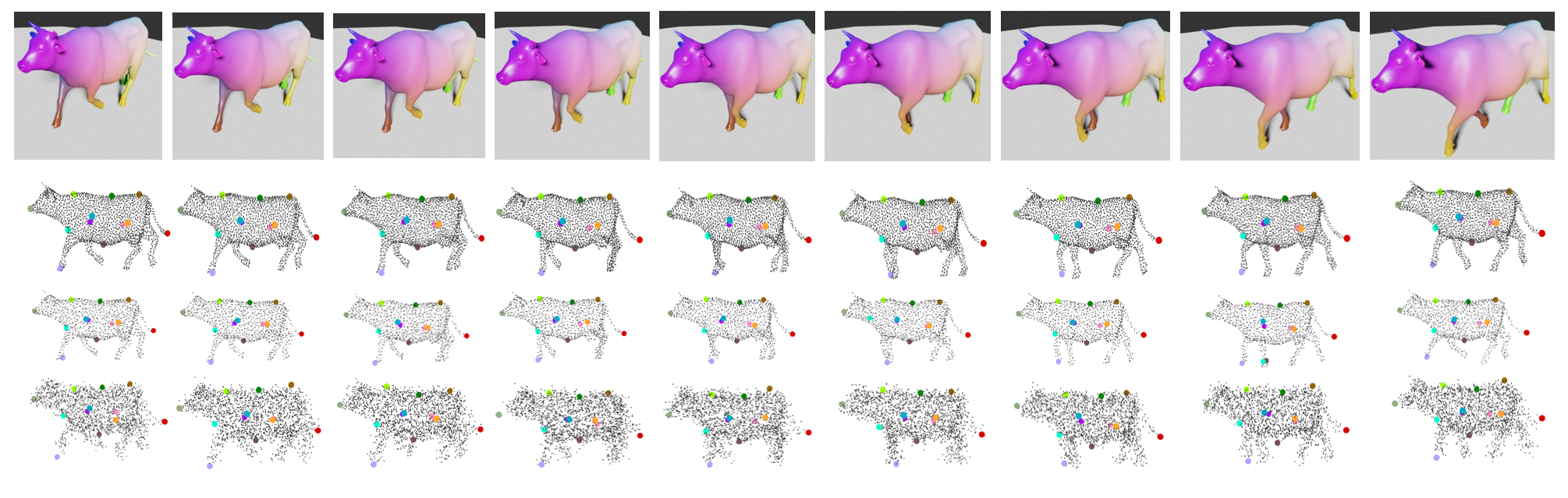  }
  \caption{Keypoints estimated by \textit{SelfGeo} on bull.  }
  \label{fig:supp_bull}
\end{figure}

\begin{figure}[]
  \centering
    \includegraphics[width=.95\linewidth]{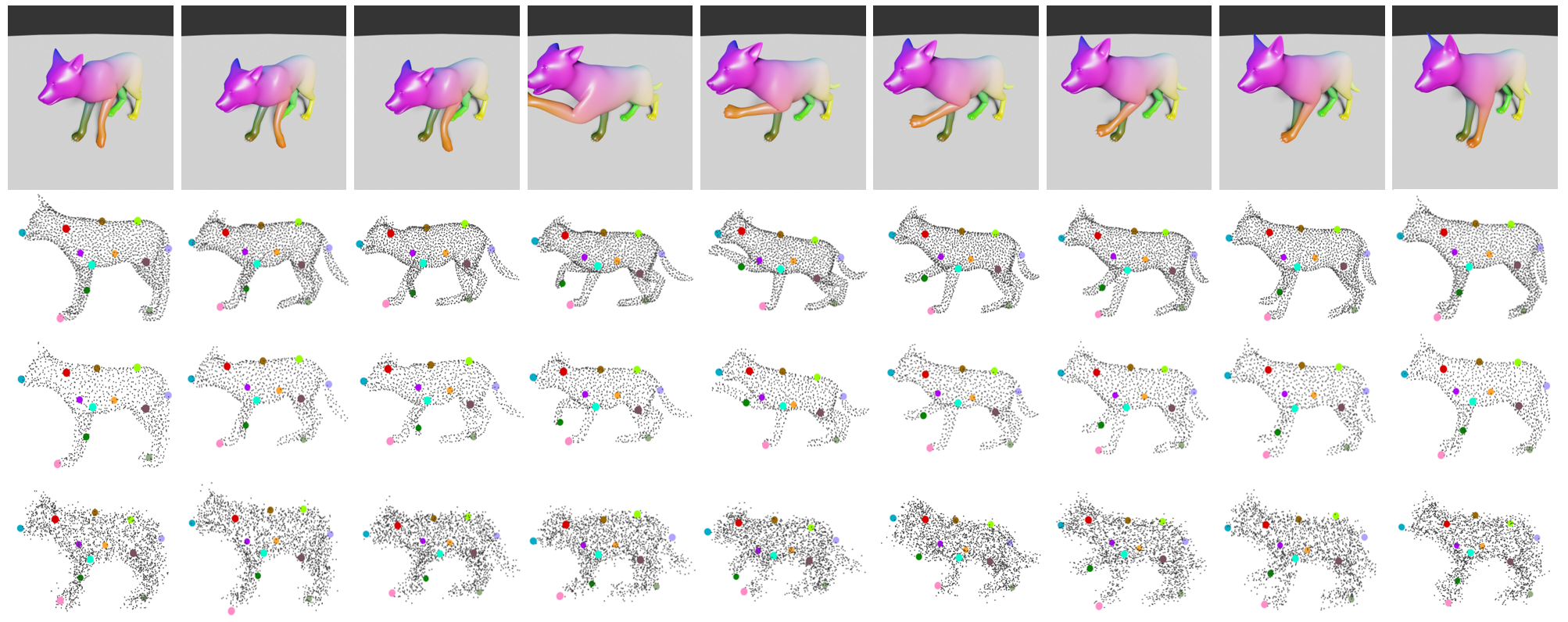}
  \caption{Keypoints estimated by \textit{SelfGeo} on dog.  }
  \label{fig:supp_dog}
\end{figure}

\begin{figure}[]
  \centering
    \includegraphics[width=.95\linewidth]{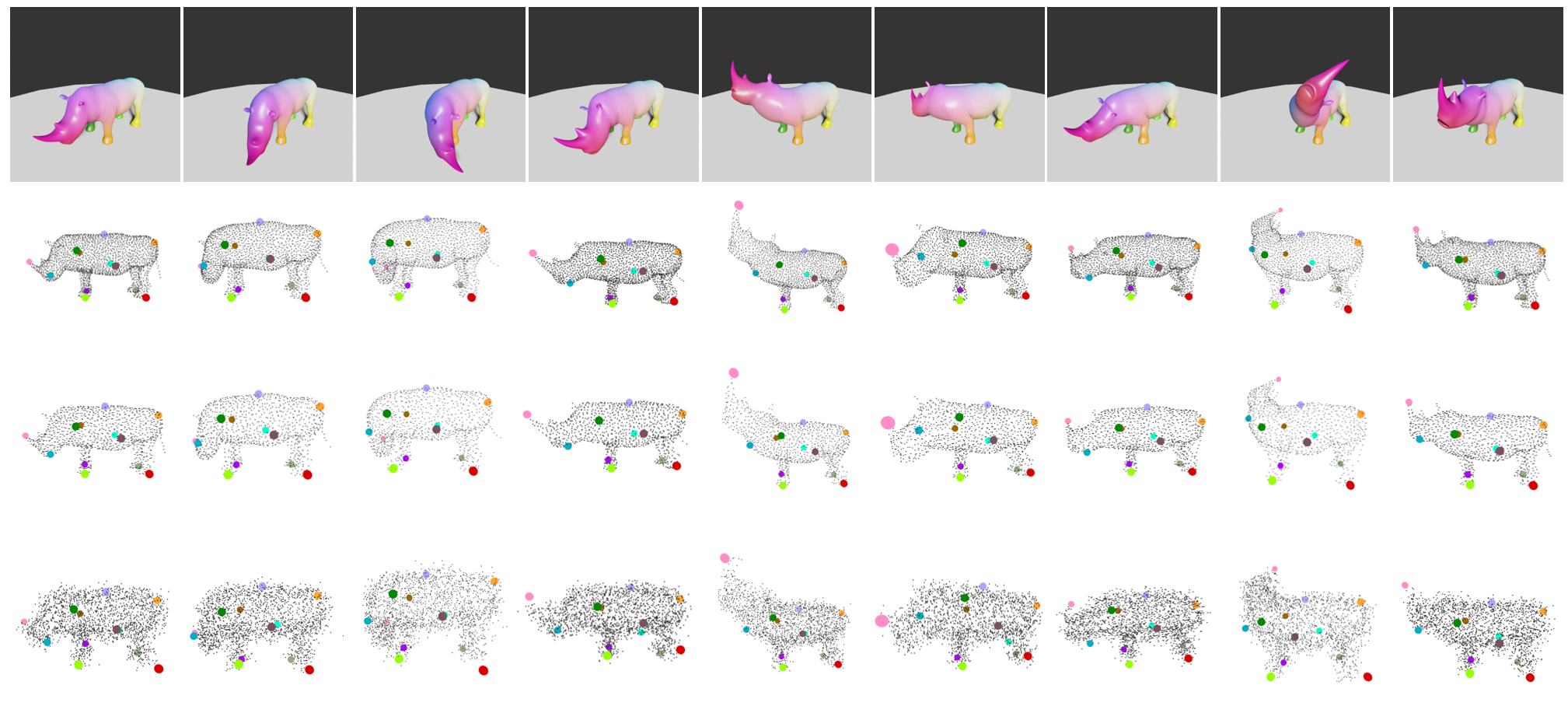}
  \caption{Keypoints estimated by \textit{SelfGeo} on rhino.  }
  \label{fig:supp_rhino}
\end{figure}

\begin{figure}[]
  \centering
    \includegraphics[width=.95\linewidth]{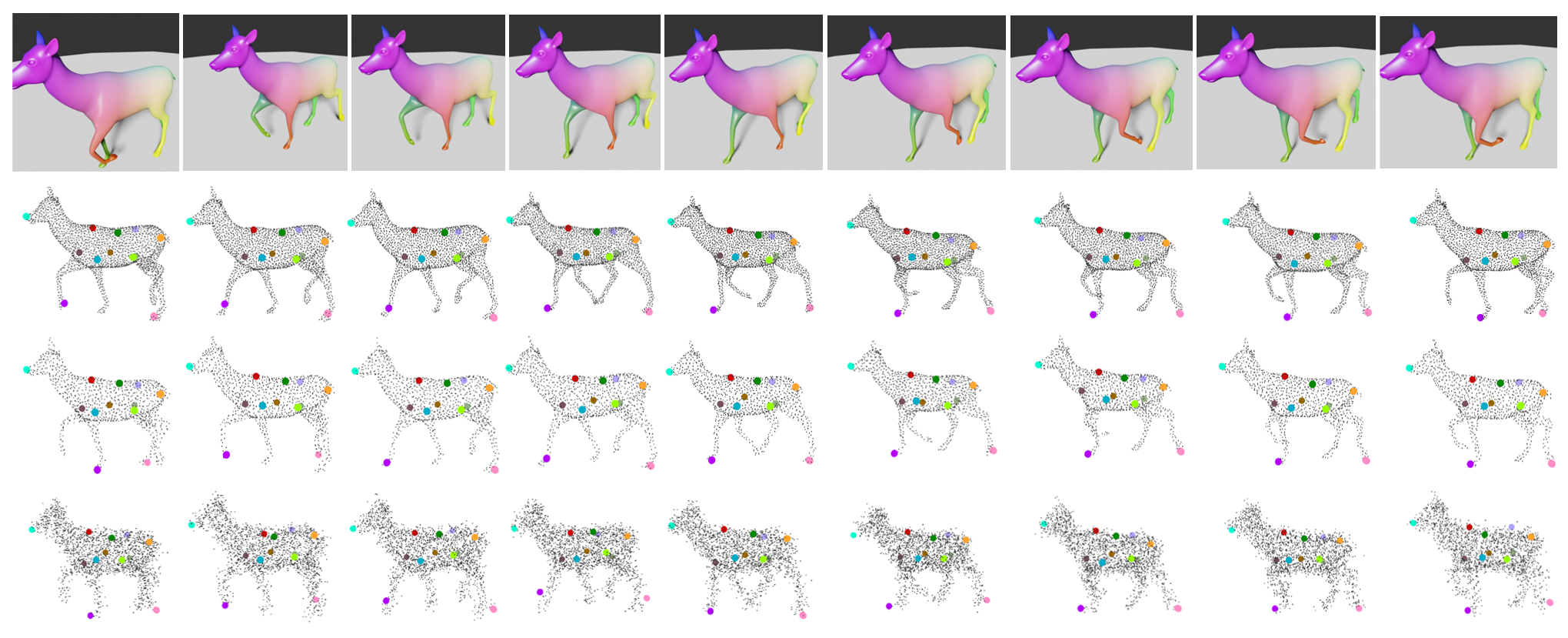}
  \caption{Keypoints estimated by \textit{SelfGeo} on Bucks.  }
  \label{fig:supp_bucks}
\end{figure}

In this section, we present qualitative results for the different categories of the Deforming Things 4D dataset. The results of bear, tiger, chicken, rabbit, deer, bull, dog, rhino, and bucks are illustrated in Fig.
~\ref{fig:supp_bear},
~\ref{fig:supp_tiger},
~\ref{fig:supp_chicken},
~\ref{fig:supp_bunny},
~\ref{fig:supp_deer},
~\ref{fig:supp_bull},
~\ref{fig:supp_dog},
~\ref{fig:supp_rhino}, and
~\ref{fig:supp_bucks}, respectively. 
Each figure contains four rows showing an animal performing an action, i.e. jumping, running, walking, attacking, etc. The first row shows images highlighting the action of the animal. These images are taken from the dataset. The second row shows the keypoints estimated by the \textit{SelfGeo} on top of the input PCDs. The third and fourth rows show the performance of the network (trained without any augmentation) on the downsampled (ratio: 2x) and noisy (variance: 0.02) PCDs, respectively. The keypoints are temporally and geometrically consistent.


\end{document}